\documentclass[10pt,twocolumn,letterpaper]{article}
\pdfoutput=1

\usepackage{url}            
\usepackage{booktabs}       
\usepackage{amsfonts}       
\usepackage{nicefrac}       
\usepackage{microtype}      
\usepackage{multirow}
\usepackage[toc,page]{appendix}
\usepackage{subcaption}
\usepackage{floatrow}
\usepackage{tablefootnote}
\usepackage[table,xcdraw]{xcolor}
\usepackage{authblk}

\usepackage{iccv} 
\definecolor{iccvblue}{rgb}{0.21,0.49,0.74}
\usepackage[pagebackref,breaklinks,colorlinks,allcolors=iccvblue]{hyperref}

\def\paperID{11809} 
\def\confName{ICCV}
\def\confYear{2025}

\title{VAGUE: Visual Contexts Clarify Ambiguous Expressions}

\author[1]{Heejeong Nam\thanks{These authors contributed equally.}}
\author[2]{Jinwoo Ahn\textsuperscript{*}}
\author[3]{Keummin Ka}
\author[3]{Jiwan Chung}
\author[3]{Youngjae Yu\thanks{Corresponding author.}}

\affil[1]{Brown University}
\affil[2]{UC Berkeley}
\affil[3]{Yonsei University}

\begin{document}
\maketitle
\let\thefootnote\relax\footnotetext{\textsuperscript{†} Authors contributed equally to this work.}

\begin{abstract}
Human communication often relies on visual cues to resolve ambiguity. While humans can intuitively integrate these cues, AI systems often find it challenging to engage in sophisticated multimodal reasoning. We introduce VAGUE, a benchmark evaluating multimodal AI systems' ability to integrate visual context for intent disambiguation. VAGUE consists of 1.6K ambiguous textual expressions, each paired with an image and multiple-choice interpretations, where the correct answer is only apparent with visual context. The dataset spans both staged, complex (Visual Commonsense Reasoning) and natural, personal (Ego4D) scenes, ensuring diversity.
Our experiments reveal that existing multimodal AI models struggle to infer the speaker's true intent. While performance consistently improves from the introduction of more visual cues, the overall accuracy remains far below human performance, highlighting a critical gap in multimodal reasoning. Analysis of failure cases demonstrates that current models fail to distinguish true intent from superficial correlations in the visual scene, indicating that they perceive images but do not effectively reason with them. We release our code and data at \href{https://hazel-heejeong-nam.github.io/vague/}{https://hazel-heejeong-nam.github.io/vague/}.

\end{abstract}
 
\section{Introduction}
\label{sec:intro} 

\begin{figure}
\centering
\includegraphics[width=\textwidth]{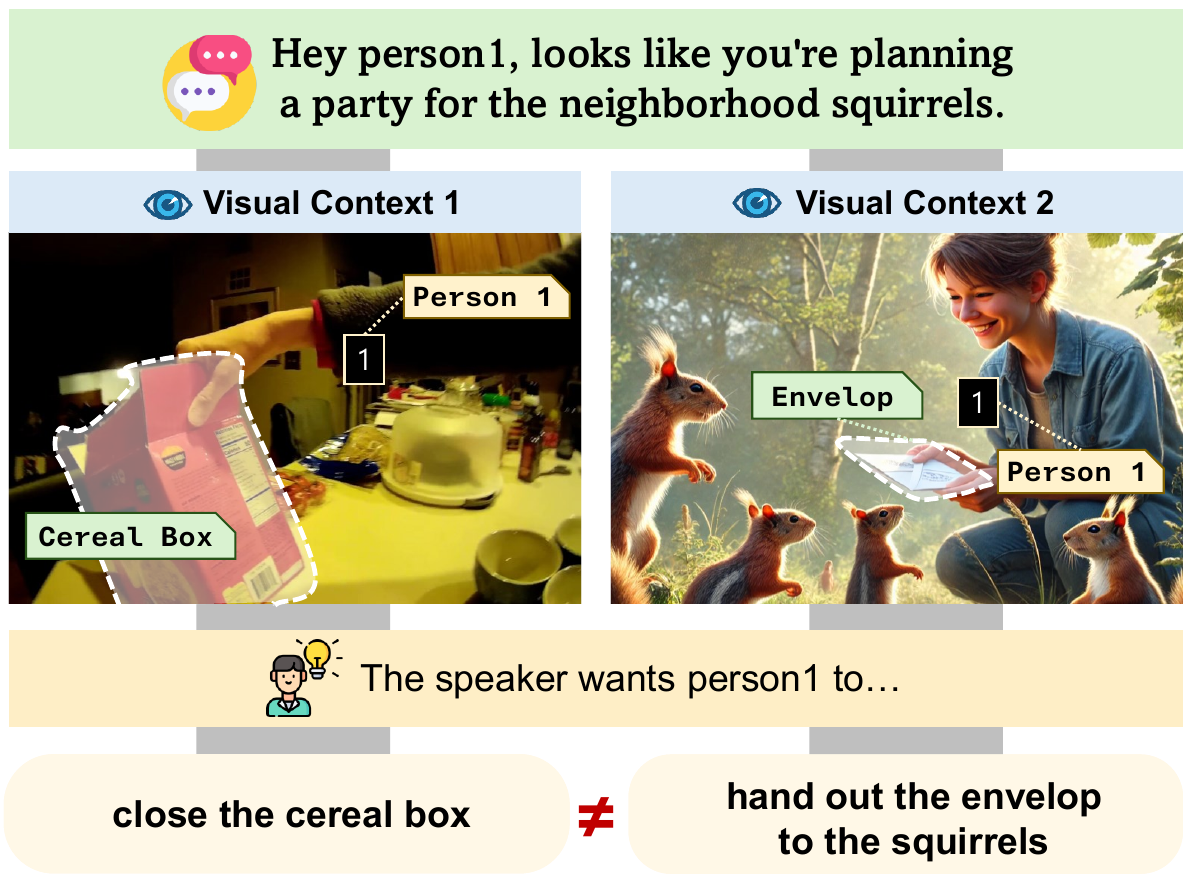}
\caption{
A motivating example demonstrating the importance of visual context in understanding intention. Without a predetermined context, a single expression can convey multiple different intentions. The textual expression and Context 1 are from our dataset, while Context 2 is generated using DALL·E 3 \cite{dalle} to help understanding.
}
\label{fig:motiv}
\end{figure}


Human communication is inherently contextual; for example, exclaiming “Hey, this is a disaster!” upon seeing a cluttered room conveys frustration or exaggeration rather than referring to an actual catastrophe. Without surrounding cues, textual dialogues can be \textit{ambiguous}, making it difficult for models to accurately capture intent and nuance.


We consider the case of \textit{visual} contextual cues. Consider~\cref{fig:motiv}, which depicts a speaker making a remark in a certain situation. Without specifying contexts introduced from visual cues, the speaker's intention can vary, thus remaining ambiguous. This implies that the visual contexts play important roles in communication, raising the question: can AI systems integrate visual cues with ambiguous dialogue to infer the speaker's intent?

We introduce \textit{\textbf{V}isual Contexts Cl\textbf{A}rify ambi\textbf{GU}ous \textbf{E}xpressions} (VAGUE), a benchmark consisting of 1.6K ambiguous textual expressions, each paired with a single image.
VAGUE aims to model diverse and natural human-to-human interactions by setting each image as the speaker’s viewpoint, where the speaker implicitly requests a certain action from a person within their field of view. We define the problem addressed through this setup as Multimodal Intention Disambiguation (MID), which involves reasoning about the most plausible request conditioned on visual context.
Each sample in VAGUE is annotated with four multiple-choice candidates, ensuring clarity and preventing multiple valid answers caused by paraphrasing or hierarchical inclusion of meaning. 
The dataset is meticulously curated to ensure visual dependency; the ground-truth candidate is only preferable when considering the visual context. VAGUE includes visual scenes from both artificial sources (Visual Commonsense Reasoning~\cite{vcr}) and real-world scenarios (Ego4D~\cite{ego4d}), capturing a broad spectrum of scene complexity and naturalness. The textual expressions in VAGUE are initially generated by GPT-4o \cite{openai2023gpt4} following instructions, then reviewed through extensive human rating and filtering to ensure naturalness and alignment with the corresponding images.

Experiments on VAGUE demonstrate that existing multimodal AI models struggle to infer a speaker's true intent in a multimodal setting. First, although models can leverage visual context—as seen by a performance progression from text-only Language Models (LMs) to pipelined Socratic Models (SMs)~\cite{zengsocratic} and ultimately to end-to-end Visual Language Models (VLMs)—their overall accuracy remains significantly lower than that of humans, indicating a failure to capture the true intent. A closer analysis of failure cases reveals that the primary source of error is the models' inability to distinguish the true intent from a superficial understanding of the visual context. In other words, even though these multimodal systems can perceive the image content, they cannot effectively use this information to reason about the speaker's true intent.

In conclusion, we introduce a benchmark that exposes the limitations of current models in integrating visual cues with intent comprehension and identifies their primary failure mode. We anticipate that VAGUE will serve as a testing ground for the development of future multimodal conversational or embodied agents—systems that combine robust visual perception with nuanced conversational reasoning to effectively respond to user requests in complex scenes.

Our contributions are threefold:
\begin{itemize}
\item VAGUE: a novel benchmark for evaluating multimodal intention disambiguation. Validated through extensive human filtering, VAGUE is designed for robust quantitative assessment by ensuring both the ambiguity of queries and the visual (in)dependency to answer candidates. 
\item Carefully curated 1,677 scene images sourced from VCR~\cite{vcr} and Ego4D~\cite{ego4d}, capturing a wide range of scene complexity, diversity, and naturalness to ensure VAGUE’s generalizability across various contexts.
\item Experimental results highlighting a critical challenge in multimodal intention disambiguation: while existing models can perceive visual cues, they fail to effectively integrate this information into reasoning to deduce the speaker's true intent.
\end{itemize}

\section{Related Work}
\label{sec:related}

\subsection{Multimodal Theory of Mind}
\label{sec:mtom}
Theory of Mind (ToM) refers to the ability to infer and reason about the intentions of others based on available information \cite{premack1978does}, where recent language models still struggle with relevant tasks \cite{cuskley2024limitations} highlighting the need for dedicated research in this area. Initially, various methods and benchmarks have been proposed in unimodal settings, relying on text-based approaches \cite{sclar-etal-2023-minding, gandhi-etal-2023-bigtom}. However, these methods often fail to capture the richness of real-world interactions, which often require integrating both linguistic and visual cues.

Moving beyond text-only contexts, recent work has incorporated visual information. MMToM \cite{mmtom} introduces a benchmark where models must process both visual and textual cues to solve question-answering tasks related to ToM. 
The BOSS dataset \cite{duan2022boss} is a multimodal dataset collected in situations where nonverbal communication is required. It is used to evaluate whether human beliefs can be inferred based on nonverbal cues during social interactions. Similarly, Chen et al. (2024) \cite{chen-etal-2024-multimodal-tom} propose a Video ToM model that leverages key video frames and transcripts, demonstrating improved reasoning on the Social-IQ 2.0 dataset \cite{siq2}. MuMA-ToM \cite{muma} further extends this direction by assessing ToM reasoning in multi-agent interactions, evaluating a model’s ability to infer human beliefs and goals based on video and text inputs. MToMnet \cite{MToMnet} introduces a ToM-based neural network that integrates contextual cues, such as scene videos and object locations, with person-specific cues, to predict human beliefs in specific scenarios. 

However, progress in multimodal ToM remains constrained not only by the scarcity of high-quality datasets \cite{chen-etal-2024-multimodal-tom} but also by the lack of explicit consideration for the ambiguity and indirectness inherent in human communication. 


\subsection{Multimodal Implicature Understanding}
\label{sec:miu}
Implicature and the ambiguity that arises from it naturally emerge in everyday human conversation, requiring pragmatic understanding \cite{sravanthi-etal-2024-pub}. Early research on implicature understanding has primarily been conducted in text-only settings \cite{28, 33, 35}, with some studies specifically focusing on figurative language and metaphor \cite{lal-bastan-2022-sbu, chakrabarty-etal-2022-flute, su-etal-2020-deepmet}. However, since the ambiguity of standalone text is inherently limited, recent studies have extended to multiple modalities. One example is multimodal sarcasm understanding (MSU). WITS \citep{23} and MOSES \cite{22} are benchmarks for sarcasm explanation, both providing the speaker's emotion and voice tone as cues. DocMSU \cite{24} is a document-level benchmark for sarcasm localization and detection. To improve MSU, EDGE \cite{26}, a graph-based approach, achieved strong performance.
UR-FUNNY \cite{urfunny} is a benchmark for multimodal humor comprehension, incorporating facial expressions and voice tones as in MSU \cite{23,22}. Hessel et al. (2023) \cite{androids} introduced a benchmark derived from a Cartoon Caption Contest, exploring humor identification and explanation. Baluja et al. (2024) \cite{baluja2024text} demonstrated that models benefit from multimodal cues in humor understanding. Memes also involve implicature, with multimodal datasets such as MemeCap \cite{hwang2023memecap} and MultiBully-Ex \cite{jha2024meme}. 

However, the cues used in multimodal implicature understanding remain simple, primarily appearing in images with a single main object or person, overlooking the importance of interactions between multiple objects and people in real-world scenarios. These limitations underscore the need for more complex cues, as addressed in VAGUE.

\begin{figure}
\centering
\includegraphics[width=\textwidth]{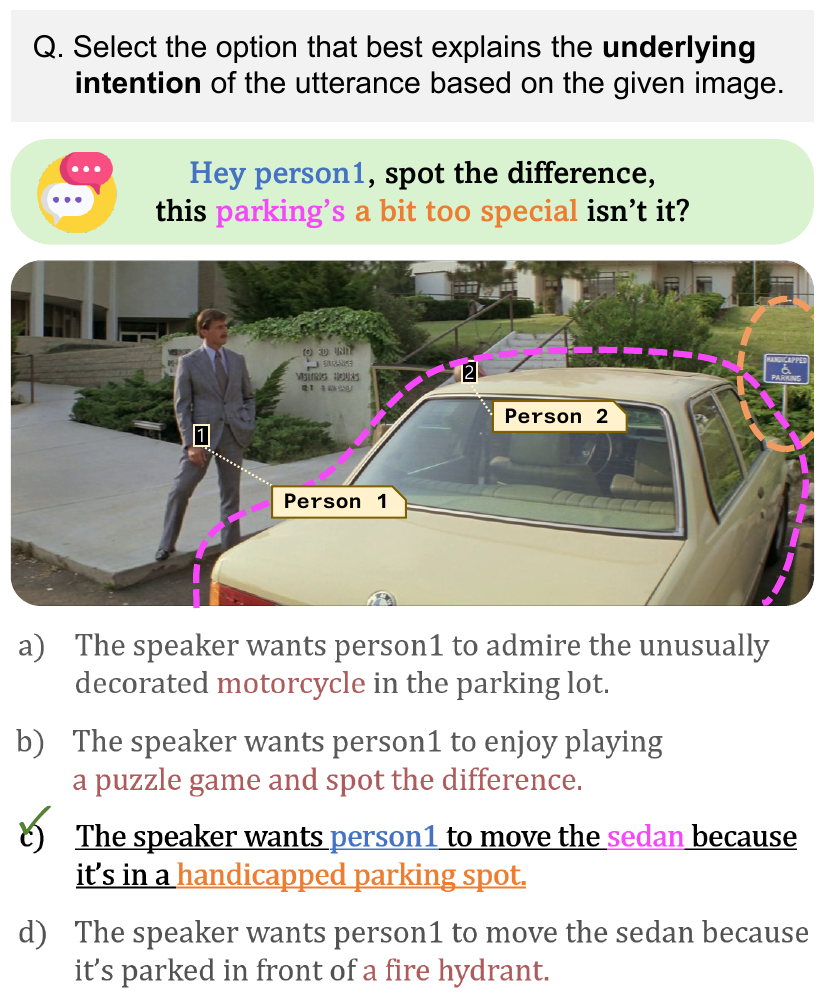}
\caption{
Description of the Multimodal Intention Disambiguation (MID) task in a Multiple-Choice Question format: Given an input image $(I)$ and an indirect expression $(p_i)$, the goal is to infer the speaker's hidden intent $(T)$ and select the most likely answer.
}
\label{fig:task}
\end{figure}

\section{Multimodal Intention Disambiguation}
\label{sec:mid}



In this section, we outline the structure and rationale behind the format of our primary task, which we term Multimodal Intention Disambiguation (MID). Then, we further specify the necessary components that form the basis of the task.

\subsection{Problem Setting}
\label{sec:prob}

Each MID problem comprises an input image \(I\), a direct text expression \(p_{d}\), and an indirect text expression \(p_{i}\). Here, the direct expression \(p_{d}\) clearly shows the underlying intention of the corresponding \(p_{i}\) and serves as an essential intermediate step of generating \(p_{i}\). 

To clarify our problem, we assume that all reasoning is confined to the depicted scene and that each expression is spoken by a human who intends for the listener to take a particular action based on the situation. The ultimate objective of the task is to interpret the hidden intention \(T\) effectively by leveraging the contextual cues within the image.

To evaluate how well models capture such intentions, we adopt a multiple-choice (MCQ) format as the primary setup, as shown in \cref{fig:task}. This decision reflects the fact that certain prompts can lead to multiple plausible outcomes, driven by hierarchical relations (e.g., pick up the \textit{chips - snack - food}) or by the inherent uncertainty of what action best satisfies the speaker’s goal (e.g., an indirect prompt complaining about darkness could be addressed by either turning on a light or opening curtains). Exploring all possible valid interpretations is labor-intensive and often infeasible. Consequently, each MID instance is presented as four distinct options, one correct and three intentionally designed to be incorrect for different reasons (see \cref{subsubsec:mcq}), challenging models in both linguistic and visual reasoning.

Formally, let \(C\) be the set of all multiple-choice options \(c_n\). Given an image \(I\) and an indirect prompt \(p_{i}\), the task is to select the most valid interpretation of \(p_{i}\) from the predefined options, conditioned on the visual context in \(I\). We define this task as follows:

\begin{equation}
T(I, p_{i}) := \operatorname{argmax}_{c_n \in C} \, \Pr(c_n \mid I, p_{i}).
\label{eq:intention}
\end{equation}


\subsection{Direct and Indirect Expressions}
\label{sec:promptdef}
By the design of our task, curating effective input prompts \(p_d\) and \(p_i\) is crucial for ensuring accurate interpretation. What makes a \textit{good} prompt, though? In this section, we define and explain the criteria that both direct and indirect expressions must satisfy. For more details on good and bad examples for each criterion, please refer to our \cref{app:textdef_example}.

\subsubsection{Directness: Relevance and Solvability}
\label{sec:directness}
\paragraph{Relevance}
The direct prompt is an utterance from the speaker that explicitly conveys its intended meaning without ambiguity. However, it is equally important that this intention aligns with the visual context of the scene.
For example, a direct prompt \( p_d \) such as “Hey person1, I want you to stop the fireworks” clearly expresses its intended action. However, if the corresponding image, as shown in \cref{fig:task}, contains no elements related to fireworks, the prompt is misaligned with the scene.
Thus, a direct prompt must not only reveal its intention but also maintain relevance to the image. In the context of our task, \textbf{relevance} is determined by whether a human can reasonably establish a connection between the prompt and the depicted scene.

\vspace{-8pt}

\paragraph{Solvability}
Relevance alone does not guarantee that a prompt is useful. As outlined in \cref{sec:prob}, the prompt \( p_d \) must explicitly request an action that the listener can reasonably perform. This introduces \textbf{solvability}, which requires that the prompt present a clear and actionable problem.
A solvable prompt defines a specific issue that can be addressed independently, ensuring that the listener is not left with multiple competing actions to choose from.


\begin{figure*}[ht]
\centering{\includegraphics[width=\textwidth]{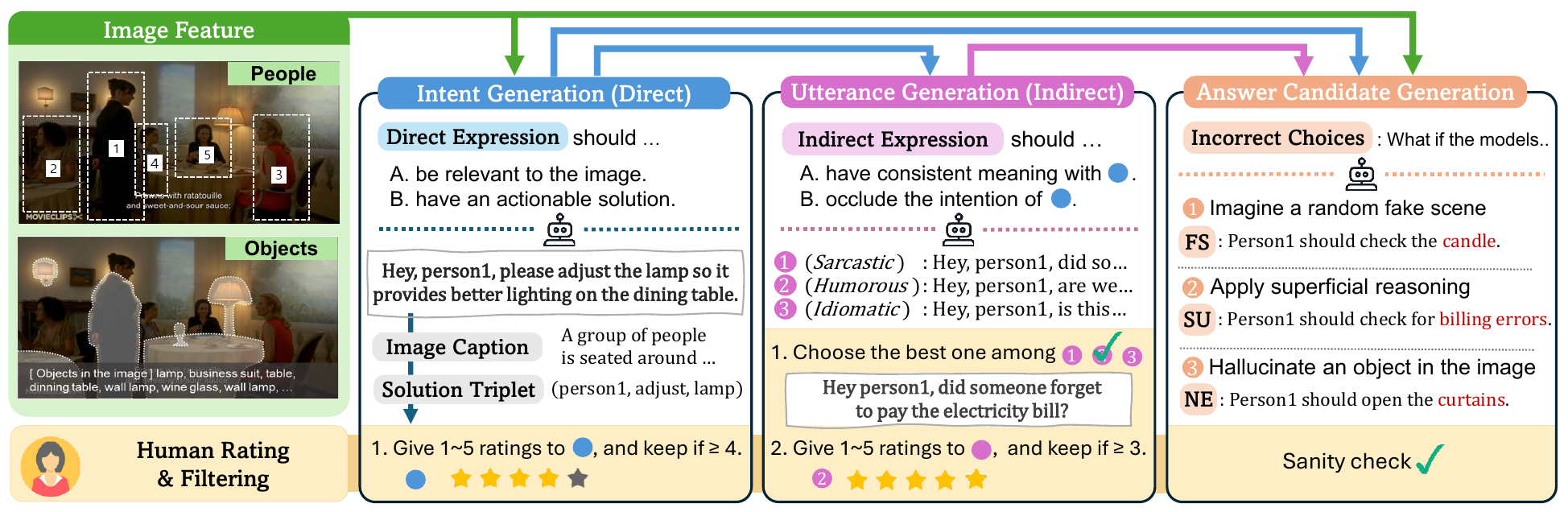}}
\caption{Overview of the data generation process. Based on human-defined criteria and instructions, GPT-4o \cite{openai2023gpt4} generates initial data, which are then rated and filtered by humans to ensure quality. Since generating high-quality indirect expressions from raw images is challenging, the process follows these steps: generating a direct expression and intention from the image, creating an indirect expression using information from the previous step, and producing answer candidates based on all the information gathered so far. In the answer candidates, FS stands for Fake Scene Understanding, SU stands for Superficial Understanding, and NE stands for Nonexistent Entity.
FS evaluates global hallucination, where the model misinterprets visual context from an entirely different image (e.g., \textit{an outdoor camping scene with a candle}), while NE addresses local hallucination, where only a specific object is replaced with a fabricated one (e.g., \textit{curtains}).
}
\label{fig:dgp}
\end{figure*}

\subsubsection{Indirectness: Consistency and Ambiguity}
\label{sec:indirectness}

\paragraph{Consistency} Indirect prompts are designed to obscure their true intention, but they must still convey the same underlying intention as their direct counterpart—essentially requesting the same solution. Since indirect prompts are derived from direct prompts, consistency serves as a key criterion. We define a direct prompt \( p_d \) and an indirect prompt \( p_i \) as consistent if their interpretations could \textit{potentially} align in intention. The term ``potentially" is used because indirect prompts, by nature, may have multiple valid interpretations. For instance, in the earlier ``this is a disaster!" example, it conveys distress but allows for multiple reasonable responses, such as cleaning the room or providing reassurance.

\vspace{-8pt}

\paragraph{Ambiguity}
If an indirect prompt is entirely consistent with its direct counterpart without introducing any additional complexity, it becomes indistinguishable from a direct prompt. Therefore, an indirect prompt should conceal its underlying intention, which we define as ambiguity. The key principle behind this criterion is that neither the specific action required nor the key entity involved should be explicitly or implicitly mentioned within the prompt. Once these two elements are concealed, further refinements—such as adjusting the tone to be more indirect, sarcastic, or humorous—can enhance the overall nuance and difficulty of interpretation.

\section{VAGUE Benchmark Construction}
\label{sec:benchmark}


VAGUE is a novel benchmark that extends single-modal or simple multimodal ambiguity to more realistic domains and evaluates whether concurrent vision-language models can perform human-like reasoning with complex visual contexts. 
It comprises 1,677 images, with 1,144 sourced from the VCR \cite{vcr} dataset and 533 from Ego4D \cite{ego4d}, covering diverse contextual scenarios as well as real-world human interactions. On average, VAGUE contains seven objects and four people per image. Each image is paired with a direct expression $p_d$, an indirect expression $p_i$, and four multiple-choice answers, along with relevant meta-information. All textual components are generated using GPT-4o, then refined through extensive human rating, selection, and filtering, ensuring a carefully curated benchmark dataset for testing and advancing multimodal reasoning. We provide detailed benchmark statistics and a diversity analysis in \cref{app:datastat}.

\subsection{Visual Data Curation}
\label{subsec:imgproc}

\subsubsection{Sampling}
\label{subsubsec:sampling}
\paragraph{VCR \cite{vcr}} The VCR dataset consists of 110K movie scenes sourced from the Large Scale Movie Description Challenge \cite{LSMDC} and YouTube clips. These images are curated based on an “interestingness” criterion \cite{vcr}, ensuring the presence of at least two people, which promotes interactive scenarios. To prevent redundancy in our dataset, we sample 10K images while carefully avoiding neighboring frames, as adjacent frames exhibit minimal variation. This selection process preserves the contextual diversity of the dataset while maintaining its focus on complex, multi-entity interactions.

\vspace{-8pt}

\paragraph{Ego4D \cite{ego4d}} While VCR provides a wide range of contextual diversity, it often includes artificially composed settings that may not fully capture real-world interactions. To address this, we integrate frames from the Ego4D dataset, which offers a more naturalistic depiction of human interactions. We specifically leverage the AV (Audio-Visual), which indicate conversational exchanges between individuals, to ensure the presence of people in the selected frames. Similar to VCR, we avoid neighboring frames to maintain diversity and filter out heavily blurry images to enhance data quality. This process results in 888 candidate images from 94 videos, which serve as the basis for further text processing.

\subsubsection{Object Extraction}  
\label{subsubsec:ram}
To ensure the complexity of the visual information, we extract a list of physical objects present in each image using a tagging model, RAM \cite{ram}. This step allows us to easily identify scenes with sufficient visual detail. In the case of VCR, many scenes are relatively simple, often containing only a few objects. Therefore, we sort VCR images by the number of detected objects and retained the top 4,000 as candidates for text processing, ensuring that our benchmark primarily consists of rich visual cues in contextually diverse scenes. Please refer to \cref{app:ram} for more details.

\subsubsection{Person Indicator} 
\label{subsubsec:person}
In our task, we assume that the speaker is outside the scene, viewing the image and talking to a person in it. However, identifying the addressee is not always straightforward, as images mostly contain multiple individuals. While grounding the specific person referenced in the utterance could introduce additional complexity, it is not the primary focus of our evaluation. To clarify the listener, we assign an indicator tag to each person in the image as shown in \cref{fig:dgp}. For VCR, we use their existing annotations \cite{vcr}, while for Ego4D, where bounding boxes are not fully available, we employ YOLOv11 \cite{yolo} to detect and annotate humans. While this provides a straightforward method for grounding the target person, it requires models to perform basic Optical Character Recognition (OCR). Therefore, we conduct experiments to evaluate the OCR capabilities of the models used in our study. See \cref{app:ocr} for details.

\subsection{Multimodal Expression Synthesis}
\label{subsec:text}
As shown in \cref{fig:dgp}, VAGUE's textual expressions are generated first by the model and undergo an extensive process of human rating and filtering. We use GPT-4o \cite{openai2023gpt4} for all text processing. The instructions used for generating direct $(p_d)$ and indirect $(p_i)$ expressions are provided in \cref{app:ins_text}, while the instructions for generating answer candidates for multiple-choice questions are detailed in \cref{app:ins_mcq}.

\subsubsection{Direct Expressions}
\label{subsubsec:d_gen}
The direct expression \( p_{d} \) serves as a crucial foundation for crafting the indirect one \( p_{i} \), as both share the same underlying intention. To ensure that  \( p_{d} \) adhere to the principles of relevance and solvability discussed in \cref{sec:directness}, we generate \( p_{d} \) conditioned on the input image \( I \) and a task prompt that explicitly defines these criteria.
During the generation process, we also instruct the model to output a solution triplet in the format: (subject, action, object). Since direct expressions explicitly state their intentions, extracting each component of the solution triplet is straightforward. To maintain consistency with the visual context, the "object" in the triplet is restricted to physical objects we extracted in \cref{subsubsec:ram}.

After generating \( p_{d} \) for all candidate images, human raters evaluate each prompt based on relevance and solvability, assigning scores from 1 to 5. Only those prompts that receive a rating of 4 or 5 are retained for further use. The detailed rating criteria for human verification and an example of the rating process are provided in \cref{fig:direct_rating}.

\subsubsection{Indirect Expressions}
\label{subsubsec:ind_gen}

To ensure that the indirect expressions \( p_{i} \) maintain both ambiguity and fluency while aligning with the true intent $T$, we adopt a two-stage process: \textbf{proposal} and \textbf{selection}. 

In the initial step, the model is prompted to generate three distinct candidate options. Each candidate follows the criteria outlined in Section \ref{sec:indirectness}, but with explicit instructions to incorporate different linguistic strategies: sarcasm, humor, and meme/idiomatic expressions, respectively. This approach ensures diversity in the generated responses.  
In the second step, human annotators evaluate the three candidates and select the one that best aligns with the intended indirectness. The selected prompt is then rated on a scale from 1 to 5 based on more specific criteria. Only those prompts that receive a score of 3 or higher are retained for use in the dataset.  
The detailed rating criteria for human verification and an example of rating process are provided in \cref{fig:indirect_rating}

\subsubsection{Counterfactual Choices}\label{subsubsec:mcq}

Generating high-quality counterfactual choices is crucial. To enable detailed analysis of model weaknesses in multimodal intent disambiguation, we design interpretable counterfactual choices that provide more plausible alternatives.

\vspace{-8pt}

\paragraph{Fake Scene Understanding}
The first counterfactual choice is an interpretation that could arise when the model largely misinterprets the image. This process is conducted in two steps. In the first step, a fake caption is generated by assuming an imaginary scene that can be aligned with the indirect expression but is inconsistent with the true intent. The caption of fake scene is then combined with the speaker's indirect statement to derive the most likely interpretation.

\vspace{-8pt}

\paragraph{Superficial Understanding}
The subsequent choice corresponds to an interpretation generated when the model fails to deeply reason about the implicit intent of the sentence and instead relies on surface-level meaning. We enforce the model to focus only on the literal wording, without considering any implied or deeper meaning of the indirect sentence. These answer choices are generated alongside the indirect expression \( p_i \). During the indirect selection phase, the corresponding superficially understood choice is selected together, maintaining coherence between them.

\vspace{-8pt}

\paragraph{Nonexistent Entity}
The last choice arises when the model interprets the text correctly but fails to adequately consider the details of the image, resulting in a plausible yet incorrect choice. This resembles the correct answer in structure, but replaces the key object in the solution with one that does not exist in the image. To prevent the task from becoming too easy by generating highly irrelevant objects, we constrain the substituted object to one that, while absent from the image, is highly expected to be present in the scene and could replace the original entity. To identify such entities, the model is provided with the image as input to choose objects that align with the scene’s context. This method ensures that the counterfactual choice leverages the expected coherence between the scene and its potential entities while rigorously testing the model’s attention to visual details.

\begin{table*}[th!]
   \tabcolsep=0.44cm
    \renewcommand{\arraystretch}{0.6}
    \small
\begin{tabular}{lcccccc}
\toprule
\multirow{3}{*}{\large Model} & \multicolumn{3}{c}{VAGUE-VCR} & \multicolumn{3}{c}{VAGUE-Ego4D} \\
\cmidrule(lr){2-4} \cmidrule(lr){5-7}
& LM & SM & VLM & LM & SM & VLM \\
& \footnotesize(L) & \footnotesize(L+V) & \footnotesize(L+V) & \footnotesize(L) & \footnotesize(L+V) & \footnotesize(L+V) \\

\midrule

Phi3.5-Vision-Instruct (4B) 
& 26.6      & 35.3 \textcolor{green!60!black}{{\footnotesize ($\uparrow$ 8.7)}}      & 46.0 \textcolor{green!60!black}{{\footnotesize ($\uparrow$ 19.4)} }
& 22.5      & 31.1 \textcolor{green!60!black}{{\footnotesize ($\uparrow$ 8.6)}}      & 42.4 \textcolor{green!60!black}{{\footnotesize ($\uparrow$ 19.9)}}\\

\midrule

LLaVA-Onevision (7B)
& 13.1      & 29.4 \textcolor{green!60!black}{{\footnotesize ($\uparrow$ 16.3)}}     & 43.1 \textcolor{green!60!black}{{\footnotesize ($\uparrow$ 30.0)}} 
& 11.3      & 29.5 \textcolor{green!60!black}{{\footnotesize ($\uparrow$ 18.2)}}     & 43.2 \textcolor{green!60!black}{{\footnotesize ($\uparrow$ 31.9)}}\\

\midrule

Qwen2.5-VL-Instruct (7B)
& 11.1      & 25.6 \textcolor{green!60!black}{{\footnotesize ($\uparrow$ 14.5)}}     & 46.8 \textcolor{green!60!black}{{\footnotesize ($\uparrow$ 35.7)}} 
& 9.8       & 28.0 \textcolor{green!60!black}{{\footnotesize ($\uparrow$ 18.2)}}     & 48.4 \textcolor{green!60!black}{{\footnotesize ($\uparrow$ 38.6)}}\\

\midrule

InternVL-2.5-MPO (8B)
& 23.0      & 48.4 \textcolor{green!60!black}{{\footnotesize ($\uparrow$ 25.4)}}     & 63.9 \textcolor{green!60!black}{{\footnotesize ($\uparrow$ 40.9)}}
& 24.2      & 54.0 \textcolor{green!60!black}{{\footnotesize ($\uparrow$ 29.8)}}     & 66.8 \textcolor{green!60!black}{{\footnotesize ($\uparrow$ 42.6)}} \\

\midrule

Idefics2 (8B) 
& 13.9      & 21.1 \textcolor{green!60!black}{{\footnotesize ($\uparrow$ 7.2)}}      & 58.7 \textcolor{green!60!black}{{\footnotesize ($\uparrow$ 44.8)}} 
& 14.8      & 18.2 \textcolor{green!60!black}{{\footnotesize ($\uparrow$ 3.4)}}      & 58.3 \textcolor{green!60!black}{{\footnotesize ($\uparrow$ 43.5)}}\\

\midrule 

LLaVA-NeXT-vicuna (13B)    
& 24.2      & 37.2 \textcolor{green!60!black}{{\footnotesize ($\uparrow$ 13)}}       & 46.4 \textcolor{green!60!black}{{\footnotesize ($\uparrow$ 22.2)}} 
& 20.3      & 34.1 \textcolor{green!60!black}{{\footnotesize ($\uparrow$ 13.8)}}     & 52.5 \textcolor{green!60!black}{{\footnotesize ($\uparrow$ 32.2)}} \\

\midrule

Ovis2 (16B)
& 21.9      & 23.8 \textcolor{green!60!black}{{\footnotesize ($\uparrow$ 1.9)}}      & 24.5 \textcolor{green!60!black}{{\footnotesize ($\uparrow$ 3.6)}} 
& 20.5      & 25.3 \textcolor{green!60!black}{{\footnotesize ($\uparrow$ 4.8)}}      & 25.7 \textcolor{green!60!black}{{\footnotesize ($\uparrow$ 5.2)}}\\ \midrule

InternVL-2.5-MPO (26B)
& 21.2      & 48.5 \textcolor{green!60!black}{{\footnotesize ($\uparrow$ 27.3)}}     & 63.7 \textcolor{green!60!black}{{\footnotesize ($\uparrow$ 42.5)}} 
& 21.8      & 55.2 \textcolor{green!60!black}{{\footnotesize ($\uparrow$ 33.4)}}     & 68.7 \textcolor{green!60!black}{{\footnotesize ($\uparrow$ 46.9)}}\\ \midrule
 InternVL-3 (38B)& 24.8& 47.2 \textcolor{green!60!black}{{($\uparrow$ 22.4)}}& 63.6 \textcolor{green!60!black}{{($\uparrow$ 38.8)}}& 18.0& 47.5 \textcolor{green!60!black}{{($\uparrow$ 29.5)}}&59.8 \textcolor{green!60!black}{{($\uparrow$ 41.8)}}
\\ \midrule
 Qwen2.5-VL-Instruct (72B)& 29.6& 55.6 \textcolor{green!60!black}{{($\uparrow$ 26.0)}}& \textbf{74.2 \textcolor{green!60!black}{{($\uparrow$ 44.6)}}}& 26.8& 59.3 \textcolor{green!60!black}{{($\uparrow$ 32.5)}}&\textbf{69.8 \textcolor{green!60!black}{{($\uparrow$ 43.0)}}}\\

\midrule

GPT-4o
& \textbf{46.4}& \textbf{69.5 \textcolor{green!60!black}{{\footnotesize ($\uparrow$ 23.1)}}}& \underline{65.1 \textcolor{green!60!black}{{\footnotesize ($\uparrow$ 18.7)}}}& \textbf{48.2}& \textbf{67.5 \textcolor{green!60!black}{{\footnotesize ($\uparrow$ 19.3)}}}& \underline{63.6 \textcolor{green!60!black}{{\footnotesize ($\uparrow$ 15.3)}}}\\

\midrule

Gemini-1.5-Pro 
& \underline{43.2}& \underline{62.4 \textcolor{green!60!black}{{\footnotesize ($\uparrow$ 19.2)}}}& 60.6 \textcolor{green!60!black}{{\footnotesize ($\uparrow$ 17.4)}} 
& \underline{40.3}& \underline{60.6 \textcolor{green!60!black}{{\footnotesize ($\uparrow$ 20.3)}}}& 60.6 \textcolor{green!60!black}{{\footnotesize ($\uparrow$ 20.3)}} \\

\bottomrule
\end{tabular}
\caption{Experiments on the Multimodal Intention Disambiguation (MID) task with varying levels of visual cues. We report the accuracy (\%) of the Multiple-Choice Question. $\uparrow$ indicates the performance gain from visual cues, i.e. increment compared to the LM setting.(L) denotes the use of language input only, while (L+V) indicates the incorporation of visual cues. The noticeable increase in accuracy across LM, SM, and VLM demonstrates that the introduction of detailed visual cues is beneficial for the task.}
\label{tab:main}%
\end{table*}%

\section{Experiments} \label{sec:exp}

\paragraph{Models}\label{subsec:models}
We use the following models in our experiments. The detailed descriptions of each model are in \cref{app:model_lst}.
\begin{itemize}
    \item Phi3.5-Vision-Instruct (4B) \cite{phi}
    \item LLaVA Onevision (7B) \cite{onevision}
    \item Qwen2.5-VL-Instruct (7B, 72B) \cite{qwen}
    \item InternVL-2.5-MPO (8B, 26B) \cite{internvl}
    \item Idefics2 (8B) \cite{idefics}
    \item LLaVA NeXT Vicuna (13B) \cite{li2024llavanextinterleavetacklingmultiimagevideo}
    \item Ovis2 (16B) \cite{ovis}
    \item GPT-4o \cite{openai2023gpt4}
    \item Gemini 1.5 Pro \cite{gemini15}
    \item InternVL-3 (38B) \cite{internvl3}
\end{itemize}

\subsection{MLLMs Benefit from Visual Cues}
\label{subsec:exp_visual_cues}

Our first objective is to assess how effectively MLLMs leverage visual cues to resolve ambiguity in utterances. To this end, we systematically control the level of detail in the visual cues provided to the models and measure their accuracy in inferring the speaker’s true intent. Performance is evaluated in both multiple-choice and free-form settings. For clarity, we primarily report multiple-choice accuracy, deferring free-form results to~\cref{app:fullres}. 

We consider three levels of visual cues:

\begin{itemize}
    \item \textit{Language Models (LMs)} receive no visual input, requiring models to rely on superficial textual priors such as common-sense knowledge of sarcasm or humor to determine intent.
    \item \textit{Socratic Models (SMs)}~\cite{zengsocratic} use text-only LMs but incorporate short image captions (up to two or three sentences) as additional input. This short generic caption may lack sufficient detail, which may be insufficient for accurately inferring intent. Each SM model generated its own image captions and used them in subsequent processing.
    \item \textit{Visual Language Models (VLMs)} receive the raw image input, enabling a more direct interpretation of visual cues.
\end{itemize}

\vspace{-8pt}

\paragraph{Results}
The results in~\cref{tab:main} indicate that MLLMs can leverage visual cues, albeit to a limited extent, since SMs and VLMs consistently outperform LMs across all evaluated models. Additionally, more detailed visual input generally improves performance, with VLMs surpassing SMs in most cases, except in the case of proprietary models. This exception is further analyzed in~\cref{subsec:exp_failure}.
Both the VCR and Ego4D subsets exhibit similar performance trends, demonstrating the generalizability of our findings across both staged and real-world scenarios. Finally, the consistently low performance of LMs further reinforces the validity of our dataset as a multimodal benchmark.

\begin{figure*}[th!]
\centering{\includegraphics[width=\textwidth]{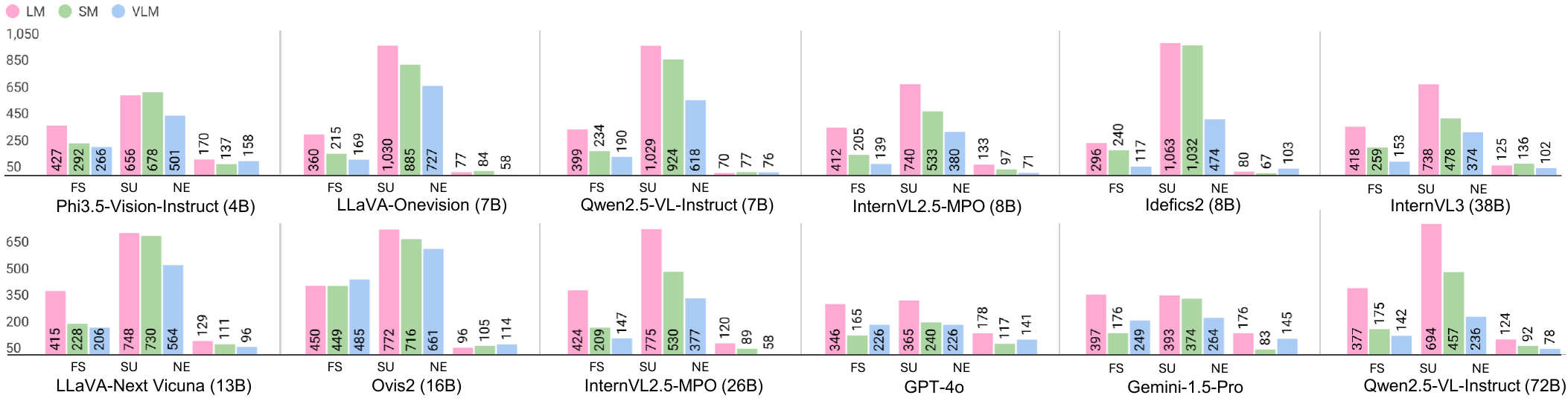}}
\caption{We present a bar plot to analyze the distribution of incorrect answer choices selected by each model. Each number represents how frequently a given choice was selected from the 1,677 items in the dataset. The counterfactual choice categories are FS (Fake Scene Understanding), SU (Superficial Understanding), and NE (Nonexistent Entity). We use distinct colors to represent LM (Language Models), SM (Socratic Models), and VLM (Visual-Language Models).}
\label{fig:barplot}
\end{figure*}

\subsection{Analysis on Failure Modes}
\label{subsec:exp_failure}

Here, we examine the ways in which models fail to infer the true intent and how these failure patterns vary with the level of visual cues provided.
As shown in~\cref{fig:dgp}, our multiple-choice questions include three distinct types of incorrect answer candidates. We assess the model’s \textit{raw visual understanding} using the Fake Scene Understanding (FS) and Nonexistent Entity (NE) candidates, which test whether the model can correctly interpret the scene without being misled by fabricated or nonexistent elements. Conversely, the Superficial Understanding (SU) candidate evaluates the model’s \textit{reasoning ability}, testing whether it can go beyond surface-level perception to infer intent. We provide the full table of failure modes in the MCQ setting in \cref{app:fullres}

\vspace{-8pt}

\paragraph{Results}
\Cref{fig:barplot} illustrates how frequently each model selects different types of incorrect answers instead of the correct intent. 
Among the error types, Superficial Understanding (SU) is the most common. This indicates that while models generally succeed in recognizing basic visual details, they often fail to \textit{reason} deeply about those visual cues to accurately infer the underlying intent of the speaker. However, proprietary models exhibit fewer SU-related errors, indicating stronger reasoning capabilities.

Moreover, stronger visual cues improve accuracy across all models and failure types. This improvement highlights the crucial role of visual conditioning in reducing both vision-based errors (FS and NE) and reasoning-related failures (SU). 
Notably, proprietary models perform better with captioned inputs (SM) than with raw images (VLM). A closer examination of~\cref{fig:barplot} reveals that this discrepancy arises from vision-based failures (FS and NE) rather than reasoning-centric failure (SU). This indicates that their captioning ability potentially allows them to obtain more sophisticated information while reducing hallucination in the images. The supporting experiments and explanations are provided in~\cref{app:sm_why_better}. 

\subsection{Comparison with human}
\label{subsec:exp_human}

\begin{table}[t]
\centering

\resizebox{\columnwidth}{!}{%
\begin{tabular}{l|c|cccc}
\toprule
Model       & Acc (\%)   & FS    &  SU   & NE    & Correct \\
\midrule
Ovis2 (16B)
& 23.0      & 119       & 171       & 18        & 92 \\
LLaVA-Onevision (7B)
& 41.0      & 43        & 183       & 10        & 164 \\
Phi3.5-Vision-Instruct (4B)
& 44.3      & 60        & 132       & 31        & 177 \\
Qwen2.5-VL-Instruct (7B)
& 47.0      & 47        & 152       & 13        & 188 \\
LLaVA-NeXT-icuna (13B)
& 48.0      & 48        & 143       & 17        & 192 \\
Idefics2 (8b)
& 57.0      & 28        & 120       & 24        & 228 \\
Gemini-1.5-Pro
& 60.3      & 60        & 73        & 26        & 241 \\
 InternVL-3 (38B)& 61.5& 38& 94& 22&246\\
InternVL-2.5-MPO (8B)
& 61.8      & 42        & 95        & 16        & 247 \\
GPT-4o
& 62.3      & 61        & 63        & 27        & 249 \\
InternVL-2.5-MPO (26B)
& 63.0      & 36        & 101       & 11        & 252 \\
 Qwen2.5-VL-Instruct (72B)& 72.3& 39& 60& 12&289\\
\midrule
\textbf{Human}       
& \textbf{94.0}      & 12        & 4         & 8         & 374 \\
\bottomrule  
\end{tabular}%
}

\caption{Performance across models and humans on a sampled set of 400 questions. The results show that humans outperform models by a margin of over 20\%.}
\label{tab:human_eval}
\end{table}

To validate our benchmark and establish an upper bound for performance, we assess human accuracy, highlighting the gap between existing models and human capability.
This evaluation follows the multiple-choice setup within the VLM setting, using a subset of 400 samples.
As shown in~\cref{tab:human_eval}, human performance reaches 94\%, demonstrating near-perfect accuracy. Although proprietary models and some large-sized models show a decent performance, a notable performance gap ($\sim$20\%) still exists compared to human evaluators. This result underscores the significant gap between AI models' multimodal reasoning capabilities and human-level understanding when inferring hidden intent, suggesting that visual perception alone, even when accurate, is insufficient without deeper cognitive integration.
This performance gap is due to the models' tendency to rely on surface-level text rather than understanding deeper visual-textual implications. 
Thus, advancing multimodal reasoning likely requires models to integrate higher-order cognitive processes, like commonsense reasoning and pragmatic understanding, into visual interpretation tasks.
Refer to~\cref{app:human_pf} for details on the selected subset and human evaluation setup.

\subsection{Chain-of-Thought Experiments}
\label{subsec:exp_cot}

\begin{table}[t]
\centering

\resizebox{\columnwidth}{!}{%
\begin{tabular}{l|l|cccc}
\toprule
\multirow{2}{*}{Model} & \multirow{2}{*}{Type} & \multirow{2}{*}{Acc (\%)} & \multicolumn{3}{c}{Incorrect count} \\
\cmidrule(lr){4-6}
    &   &   &   FS  &   SU   &  NE \\
\midrule
\multirow{4}{*}{GPT-4o} 
& SM        & 68.9      & 165   & 240   & 117 \\
& SM+CoT    & 69.5 \textcolor{green!60!black}{{\footnotesize ($\uparrow$ 0.6)}}     & 165   & 241   & 105 \\ \cmidrule(lr){2-6}
& VLM       & 64.6     & 226   & 226   & 141\\
& VLM+CoT   & 66.4 \textcolor{green!60!black}{{\footnotesize ($\uparrow$ 1.8)}}     & 162   & 156   & 85\\
\midrule
\multirow{4}{*}{Gemini-1.5-Pro} 
& SM        & 61.8      & 176   & 374   & 83 \\
& SM+CoT    & 61.0 \textcolor{red!60!black}{{\footnotesize ($\downarrow$ 0.8)}}     & 190   & 367   & 94 \\ \cmidrule(lr){2-6}
& VLM       & 60.6      & 249   & 264   & 145 \\
& VLM+CoT   & 64.4 \textcolor{green!60!black}{{\footnotesize ($\uparrow$ 3.8)}}     & 213   & 267   & 117 \\
\bottomrule
\end{tabular}
}
\caption{Result of Chain-of-Thought (CoT) experiments on proprietary models, in both SM and VLM settings. $\uparrow$ and $\downarrow$ indicate an increase and decrease in accuracy when zero-shot CoT is applied.}
\label{tab:private_accuracy-main}
\end{table}

Given the strong reasoning demands of multimodal intent deduction, we further explore the effectiveness of Chain-of-Thought (CoT) prompting~\cite{wei2022chain} in enhancing the reasoning capabilities of MLLMs.
The CoT prompt templates, provided in~\cref{fig:VLM_cot_mcq} and ~\cref{fig:SM_cot_mcq}, are designed to explicitly ground the reasoning process, reducing hallucinations.
While our main results focus on proprietary models, owing to their superior suitability for zero-shot CoT reasoning, we also present experiments and analyses for open-source models ranging from 4B to 72B in~\cref{app:cot_open}.

\vspace{-8pt}

\paragraph{Results}
As shown in~\cref{tab:private_accuracy-main}, CoT prompting improves performance for raw image inputs (VLM), while showing no clear trend and remaining at a similar level for image caption inputs (SM). One possible explanation for this discrepancy is that CoT primarily enhances reasoning by improving grounding and reducing hallucinations. Since image captions inherently contain fewer hallucinations, SMs see some benefit from CoT prompting albeit at the cost of reduced detail.
Additionally, the performance improvements observed with CoT prompting are consistent across different types of false answer candidates, suggesting a generalizable effect in enhancing reasoning quality.  

\section{Conclusion} 
We present VAGUE (Visual Contexts ClArify ambiGUous Expressions), a new benchmark aimed at assessing models' ability to interpret nuanced communication in complex multimodal scenarios. Our results show that models benefit from visual information when inferring the underlying intention of indirect expressions, as evidenced by their improved performance with increasing levels of visual cues. However, a significant disparity persists between machine capabilities and humans. To gain deeper insights into model inaccuracies, we design multiple-choice questions that explicitly address failure points, enabling a systematic and quantitative evaluation of the reasons behind performance. The primary challenge identified is the tendency of multimodal models to inadequately integrate visual cues, relying instead on the literal interpretation of textual information. This shortcoming highlights the need for further research, and we anticipate that VAGUE will open up promising avenues for developing systems capable of deeper multimodal reasoning to enhance AI's ability to engage in human-like interactions.

\section{Limitations}

First, we acknowledge that there are cultural and linguistic limitations. We incorporate sarcastic, humorous, and idiomatic implicatures in generating indirect expressions. However, since the initial data drafts are created by a model (GPT-4o \cite{openai2023gpt4}), they may reflect cultural biases present in its training data. To prevent any potential ethical issues, all human annotators are instructed to remove any content considered problematic or discriminatory during the rating and filtering process. Also, all textual expressions in VAGUE are limited to English. Therefore, we encourage future exploration of indirect expressions across diverse languages and cultures. The second limitation pertains to the dependency of certain meta-information on the quality of the parent dataset and the performance of the models utilized during the dataset generation. We observe that bounding box annotations from YOLOv11 \cite{yolo} are occasionally duplicated, leading to a reported number of people higher than actually present. Likewise, the tagging model RAM \cite{ram} sometimes misidentified objects. Therefore, we remove any corrupted instances that could undermine the integrity of the task during the human rating and filtering process.

\vfill

\newpage

{
    \small
    \bibliographystyle{ieeenat_fullname}
    \bibliography{main}
}

\newpage
\clearpage
\setcounter{page}{1}
\maketitlesupplementary
\appendix
\renewcommand{\thesection}{\Alph{section}} 
\renewcommand{\thefigure}{\Alph{section}\arabic{figure}} 
\renewcommand{\thetable}{\Alph{section}\arabic{table}} 
\setcounter{figure}{0} 
\setcounter{table}{0} 

\section{Directness \& Indirectness Examples}
\label{app:textdef_example}
\begin{figure*}[bh!]
\centering
\includegraphics[width=0.9\textwidth]{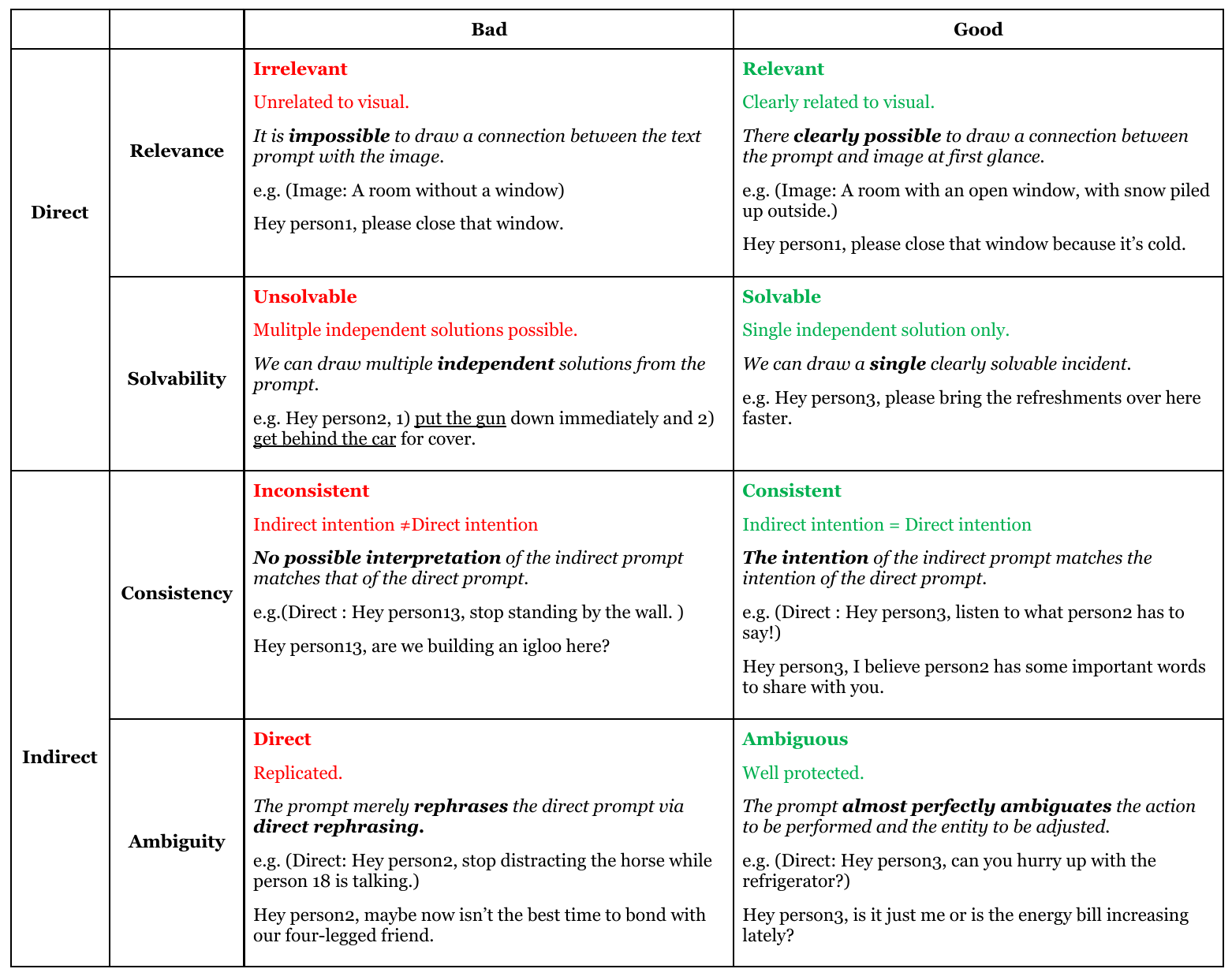}
\caption{This table presents two evaluation criteria for direct expressions and two for indirect expressions, along with descriptions and corresponding bad and good cases. The examples in the bad and good cases are derived from human ratings based on the given criteria.
}
\label{fig:directness_indirectness_examples}
\end{figure*}

\Cref{fig:directness_indirectness_examples} provides a more concrete explanation with realistic examples of the direct and indirect expressions defined in the main text. Each expression follows two key criteria, and we avoid examples like those in the ``Bad" column while prioritizing those in the ``Good" cases.

\section{VAGUE Benchmark}
This section introduces details of VAGUE Benchmark dataset. We provide examples of images and their corresponding multiple-choice questions, along with the prompts used to generate direct, indirect expression, correct understanding expression and superficial understanding expression. Additionally, we present the prompts used to generate two incorrect answer choices of multiple choices set (fake scene understanding expression, nonexistent entity expression) and the human rating criteria employed to assess the quality of direct and indirect expressions.

\subsection{Samples in VAGUE}

\begin{figure*}[t]
\centering
\includegraphics[height=21cm, keepaspectratio]{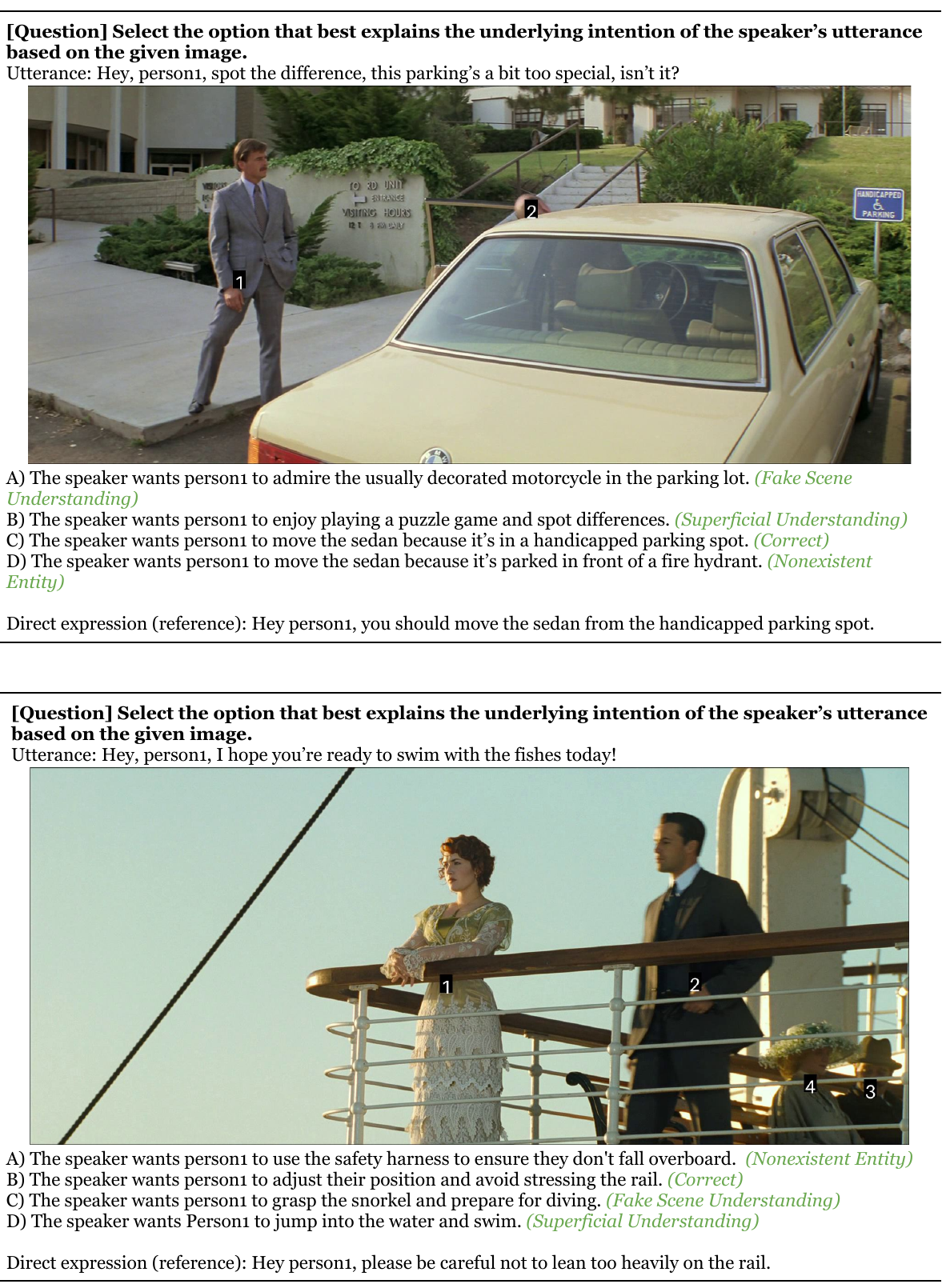}
\caption{These are examples of an image along with its corresponding generated direct expression and multiple-choice question set.
}
\label{fig:example1}
\end{figure*}

\begin{figure*}[t]
\centering
\includegraphics[height=21cm, keepaspectratio]{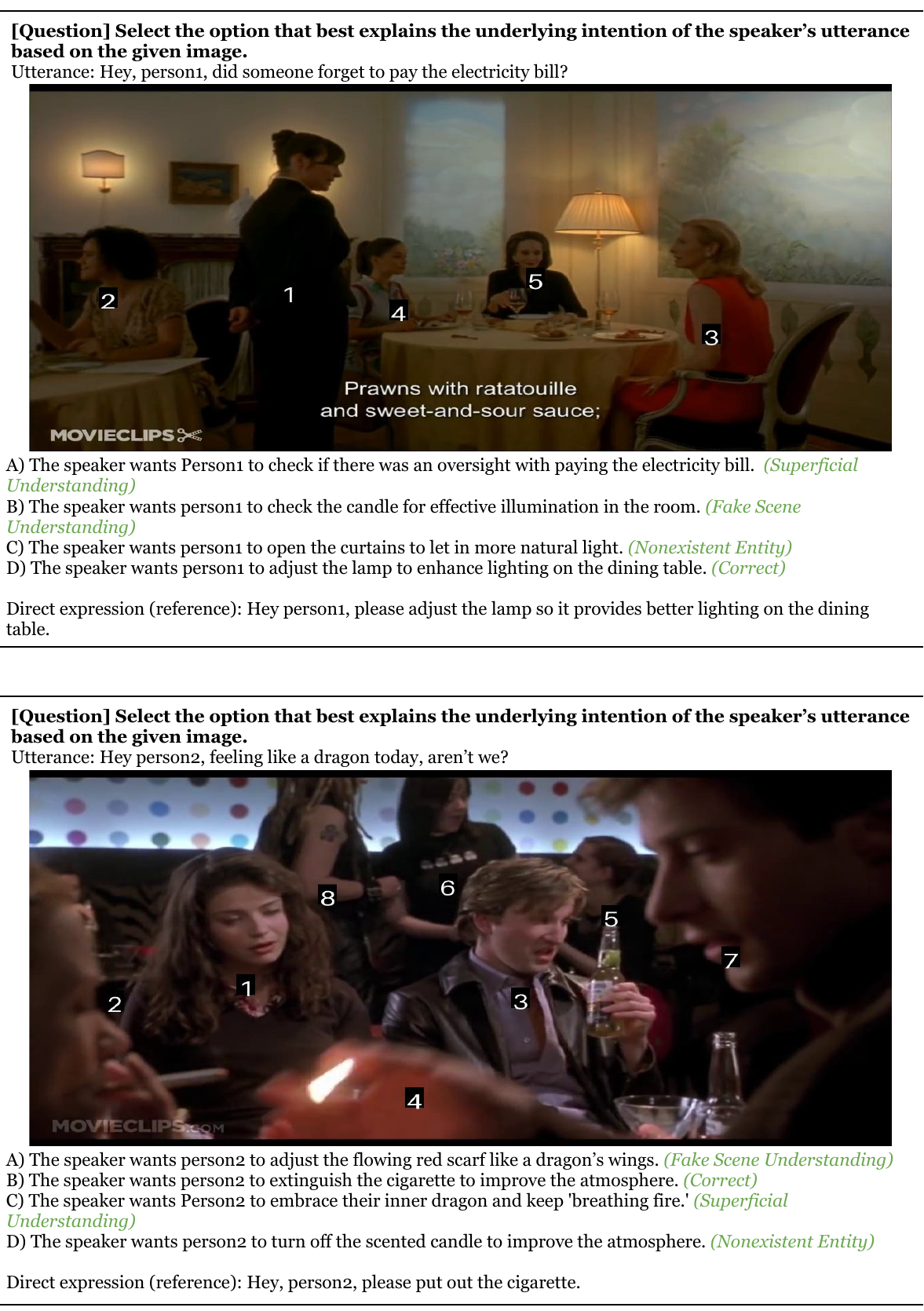}
\caption{These are examples of an image along with its corresponding generated direct expression and multiple-choice question set.
}
\label{fig:example2}
\end{figure*}

\begin{figure*}[t]
\centering
\includegraphics[height=21cm, keepaspectratio]{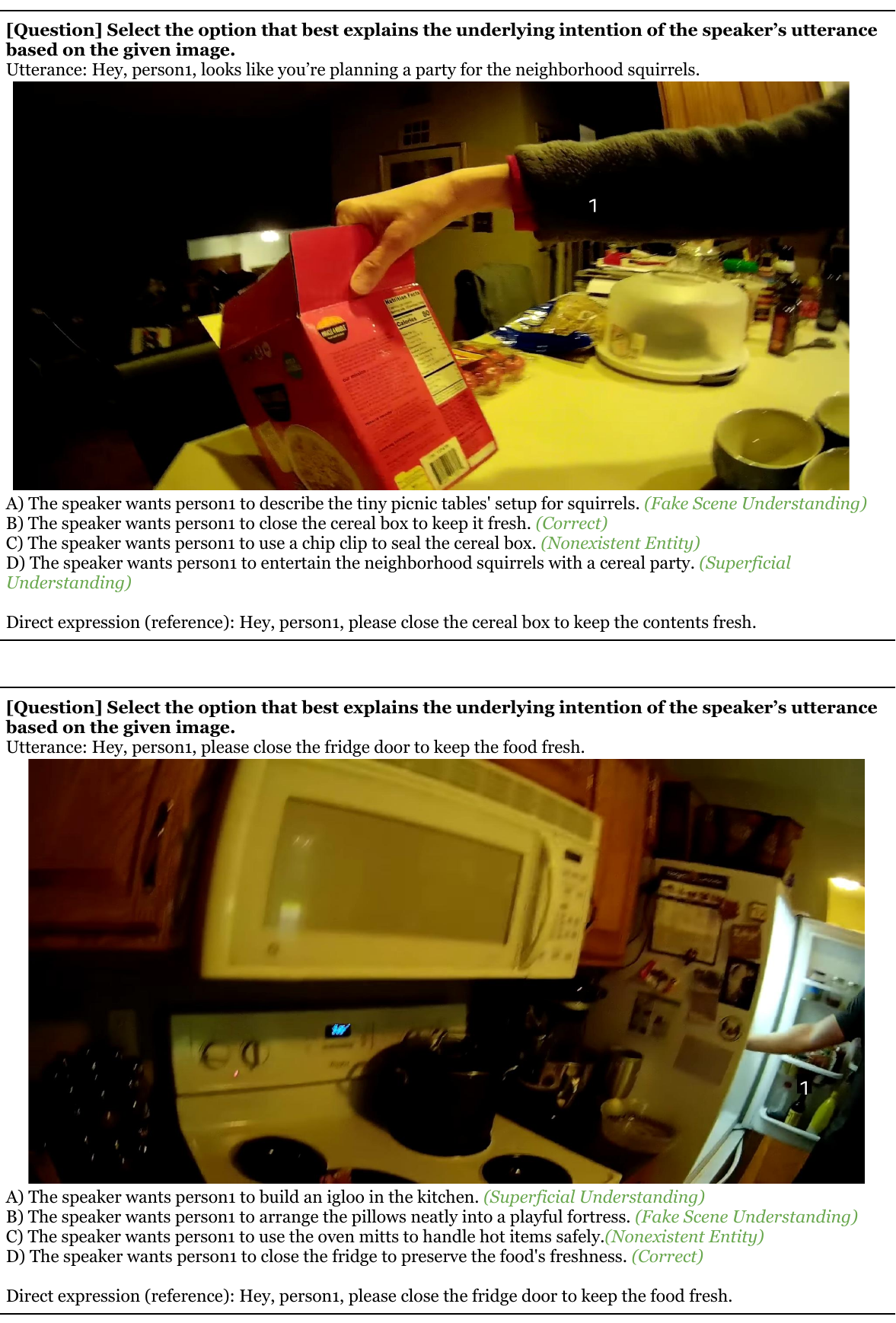}
\caption{These are examples of an image along with its corresponding generated direct expression and multiple-choice question set.
}
\label{fig:example3}
\end{figure*}

\Cref{fig:example1,fig:example2,fig:example3} illustrate six examples from our benchmark dataset. In the Visual Language Models (VLMs) setting, the model is presented with an image containing person indicator tags, a question, and the speaker’s indirect utterance, as shown in the figures, to answer a multiple-choice question. For reference, we have included the corresponding direct expression below each sample.

\subsection{Benchmark Statistics}\label{app:datastat}

\Cref{tab:dataset_statistics} illustrates the average statistics of the datasets that make up VAGUE-VCR and VAGUE-Ego4D. ``Average object counts" refers to the average number of detected objects per image, while ``Average people counts" indicates the average number of detected individuals per image. ``Average word counts" represents the average number of words in direct and indirect expressions generated for each image. Notably, the high values of ``Average object counts" and ``Average people counts" suggest that the images are not simplistic.

\begin{figure*}[t]
    \centering
    \begin{subfigure}[b]{\textwidth}
        \centering
        \includegraphics[width=\textwidth, height=0.45\textheight, keepaspectratio]{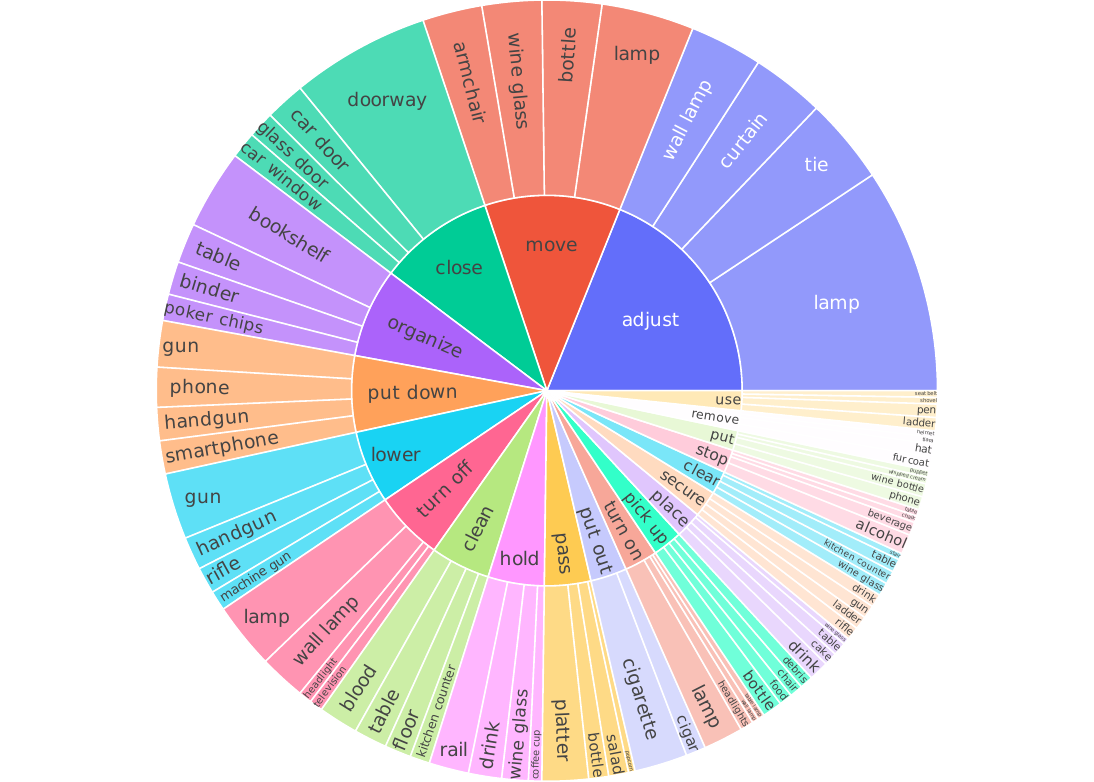}
        \caption{Diversity diagram of VAGUE-VCR}
        \label{fig:image1}
    \end{subfigure}
    \vspace{0.3cm} 
    \begin{subfigure}[b]{\textwidth}
        \centering
        \includegraphics[width=\textwidth, height=0.45\textheight, keepaspectratio]{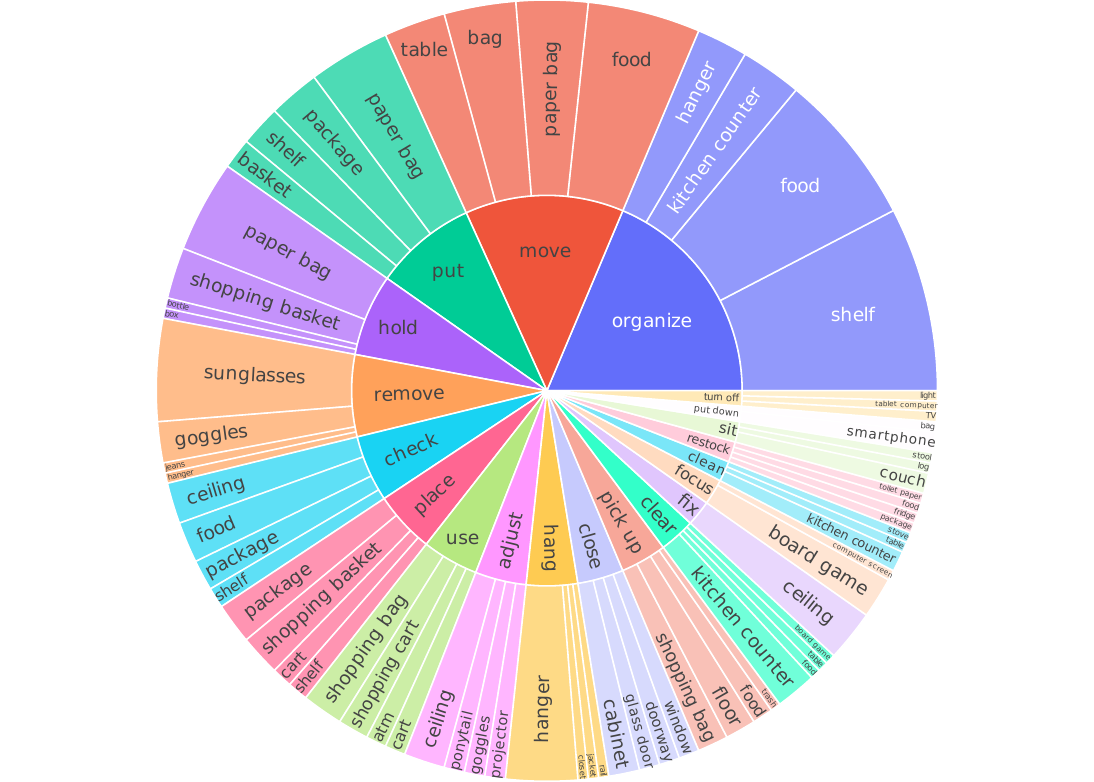}
        \caption{Diversity diagram of VAGUE-Ego4D}
        \label{fig:image2}
    \end{subfigure}
    \caption{Diversity diagrams of the 20 most frequent actions in VAGUE-VCR dataset and VAGUE-Ego4D dataset respectively.}
    \label{fig:stat_diverstiy}
\end{figure*}
\begin{table}[ht!]
\centering

\resizebox{\columnwidth}{!}{%
\begin{tabular}{l|cc}
\toprule
    & VAGUE-VCR & VAGUE-Ego4D \\
\midrule
Average object counts   & 7.4   & 6.89 \\
Average people counts   & 4.48  & 2.59 \\
Average word counts (direct)    & 9.69  & 15.9 \\
Average word counts (indirect)  & 11.52 & 12.27 \\
\bottomrule  
\end{tabular}%
}

\caption{Dataset statistics table}
\label{tab:dataset_statistics}
\end{table}

Furthermore, \cref{fig:stat_diverstiy} illustrates the diversity of intentions generated from our two parent datasets. Using the 20 most frequently occurring verbs in the solution triplets (person, action, object) of each dataset, we generate a radial diagram. Both VAGUE-VCR and VAGUE-Ego4D exhibited a comparable level of diversity, demonstrating that while not perfectly uniform, the dataset covers a wide range of contexts.

\subsection{Details of Object Extraction}
\label{app:ram}
Our prompts, both direct and indirect, are crafted as instructions that request the recipient to perform specific manipulations on an object within the scene. Such formulation of the task prompt requires the scene to have enough objects. Although the VCR \cite{vcr} dataset contains COCO \citep{coco} object tags as meta-information, COCO objects are highly limited and often fail to comprehensively capture the objects present in real-world scenes. Therefore, we process our images using the Recognize Anything Model (RAM) \citep{ram} to identify physical objects in each image. However, RAM \citep{ram} frequently generates tags for entities that are not strictly physical objects, such as places, emotions, and colors. To address this, we manually curate a list of 2,403 physical objects from the full set of 4,585 items detectable by RAM \citep{ram}. Using this refined list, we filter the initially extracted entities from the images for further usage.

\subsection{OCR Experiments for Testing Person Tags}
\label{app:ocr}
\begin{figure}[h!]
\centering
\includegraphics[width=\textwidth]{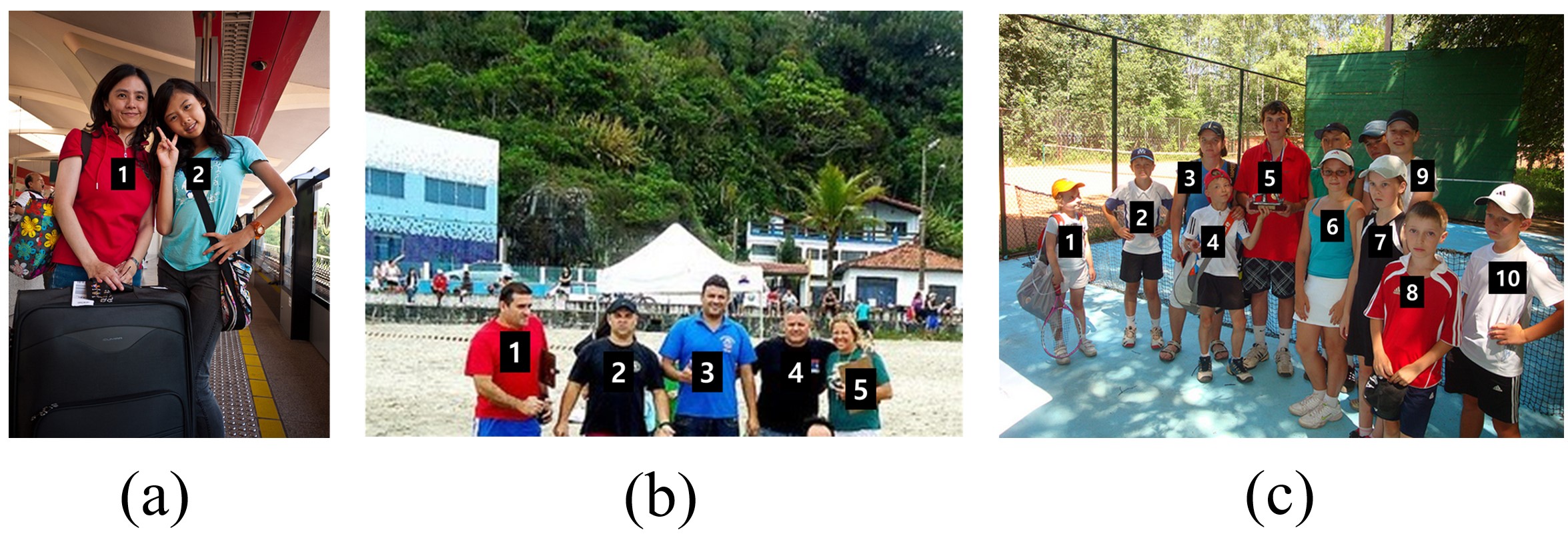}
\caption{Three images from the COCO \cite{coco} dataset which were used assessing OCR capability. We asked about the t-shirt color of person 1 in (a) and (b), and person 5 in (c). The correct answer, "red," was identified correctly by all selected models during testing.  }
\label{fig:ocr}
\end{figure}
In \cref{subsubsec:person}, we incorporate person indicators into images to distinguish individuals when interpreting prompts in various contexts. While this provides a straightforward method for grounding the target person, it requires models to perform basic Optical Character Recognition (OCR) to interpret phrases such as ``Hey person2" To evaluate this capability, we select three images from the COCO \cite{coco} dataset, each containing a different number of people. Fig. \ref{fig:ocr} presents these images, featuring two, five, and ten individuals wearing t-shirts in various colors. By asking the models to identify the t-shirt color of specific individuals, we conclude that the selected models consistently performed perfectly in recognizing person indicators as shown in \cref{tab:ocr_experiment}
\begin{table*}[h]
\ttabbox[]
  {\centering
   \tabcolsep=0.07cm
    \renewcommand{\arraystretch}{1}
    \small
\begin{tabular}{@{}cccc@{}}
\toprule
\multirow{2}{*}{model} & \multicolumn{3}{c}{response}                                                                                                 \\ \cmidrule(l){2-4} 
& (a)   & (b)   & (c) \\ 
\midrule
Phi3.5-Vision-Instruct (4B)          
&   
Person1 is wearing a red shirt.& 
Person1 is wearing a red shirt.& Person5 is wearing a red shirt.\\

LLaVA-Onevision (7B) 
& 
red&  
red&       red\\

Qwen2.5-VL-Instruct (7B)
& Person1 is wearing a red shirt.
& Person1 is wearing a red shirt. 
& Person5 is wearing a red shirt.            \\

InternVL-2.5-MPO (8B)   
& Person 1 is wearing a red shirt. 
& Person 1 is wearing a red shirt. 
& Person 5 is wearing a red shirt.             \\

Idefics2 (8B)    
& Red.        
& Red.        
& Red.            \\

LLaVA-NeXT-vicuna (13B)  
& Person 1 is wearing a red t-shirt.  
& Person 1's t-shirt is red.          
& Person 5 is wearing a red t-shirt.               \\

Ovis2 (16B)        
& Person1 is wearing a red shirt.        
& Person1 is wearing a red shirt.       
& Person5 is wearing a red shirt. \\

InternVL-2.5-MPO (26B)                
& Person 1 is wearing a red shirt.
& Person 1 is wearing a red shirt.
& Person 5 is wearing a red shirt.              \\
 InternVL-3 (38B)& Person 1 is wearing a red shirt.& Person 1 is wearing a red shirt.&Person 5 is wearing a red shirt.              \\
 Qwen2.5-VL-Instruct (72B)& Person 1 is wearing a red shirt.& Person 1 is wearing a red shirt.&Person5 is wearing a red shirt.            \\ 

GPT-4o              
& 
Person 1 is wearing a red shirt.& 
Person 1 is wearing a red shirt.&             Person 5 is wearing a red shirt.\\

Gemini-1.5-Pro        
& Person 1 is wearing a red t-shirt. 
& Person 1 is wearing a red t-shirt. 
& Person 5 is wearing a red t-shirt.             \\

\bottomrule
\end{tabular}
      \vspace{-7pt}
      }%
    {\caption{}\label{tab:ocr_experiment}}%
\end{table*}%

\subsection{Model Instruction for Direct and Indirect prompts}
\label{app:ins_text}
\begin{figure*}[t]
\centering
\includegraphics[width=\textwidth]{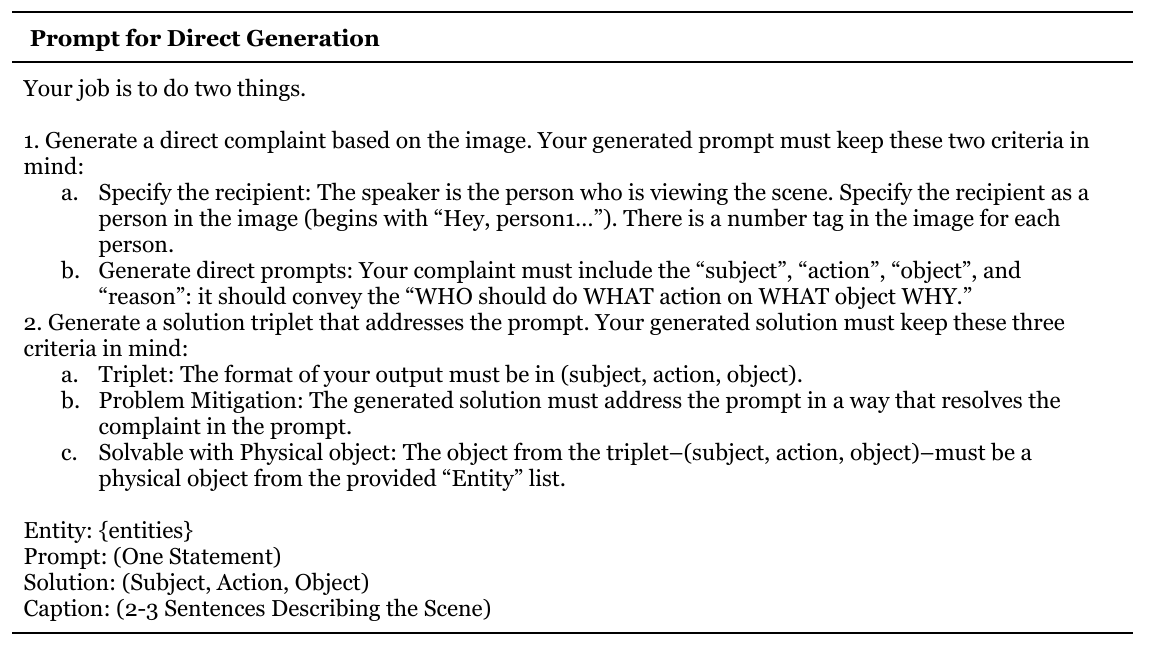}
\caption{This prompt selects one of the list of entities and generates a direct request to a person in the image. It also generates a triple solution and generates a caption for the scene.
}
\label{fig:direct_generation}
\end{figure*}

\begin{figure*}[t]
\centering
\includegraphics[width=\textwidth]{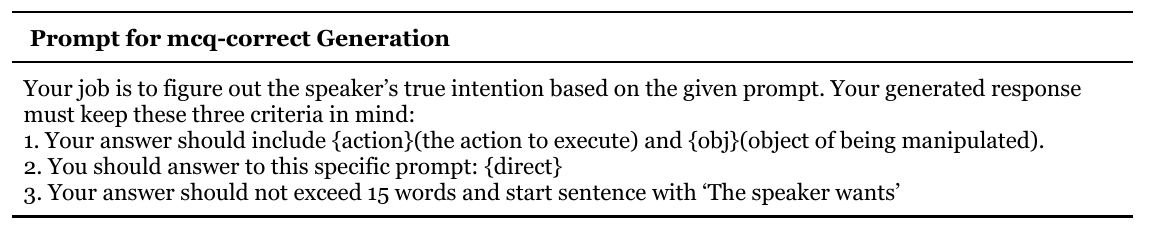}
\caption{This prompt takes an action, object, and direct expression as input and outputs the true intention behind the direct expression.
}
\label{fig:correct_generation}
\end{figure*}

\begin{figure*}[t]
\centering
\includegraphics[width=\textwidth]{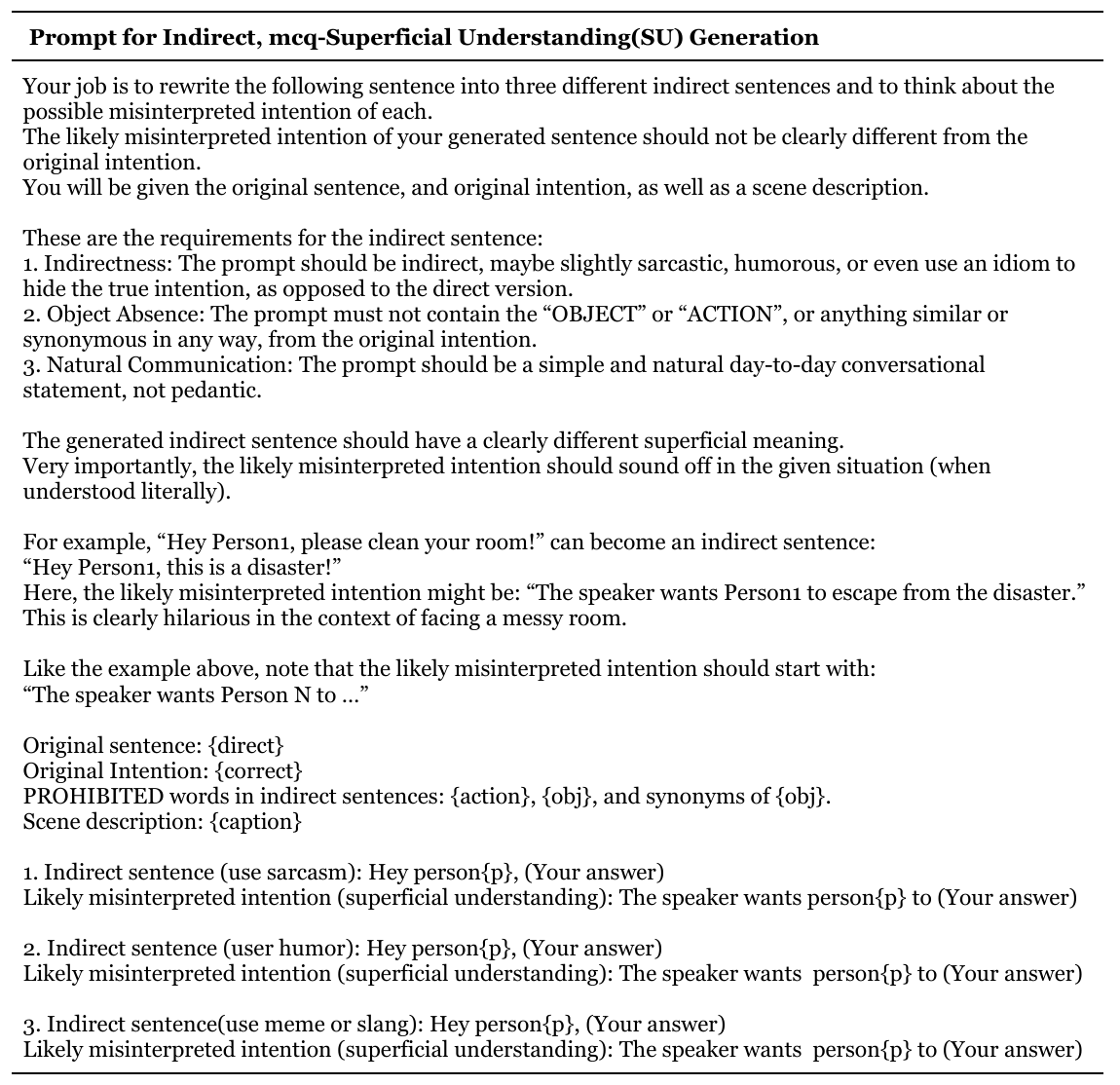}
\caption{This prompt takes a direct expression, true intention, and caption as input to generate an indirect expression that conveys the intended meaning. Additionally, it produces a superficially interpreted version of the indirect expression.
}
\label{fig:indirect_surface_generation}
\end{figure*}

\cref{fig:direct_generation} is the prompt used to generate direct expressions. Also, \cref{fig:correct_generation} is a prompt that generates the correct answer for multiple choice by understanding the intention based on this direct expression(Correct). \cref{fig:indirect_surface_generation} is the prompt that generates indirect expressions based on direct expressions and their intended meaning. Additionally, it produces superficially misinterpreted intentions (SU). 

\subsection{Human Rating for Direct and Indirect prompts}
\label{app:humanrate}

\begin{figure*}[t]
\centering
\includegraphics[width=\textwidth]{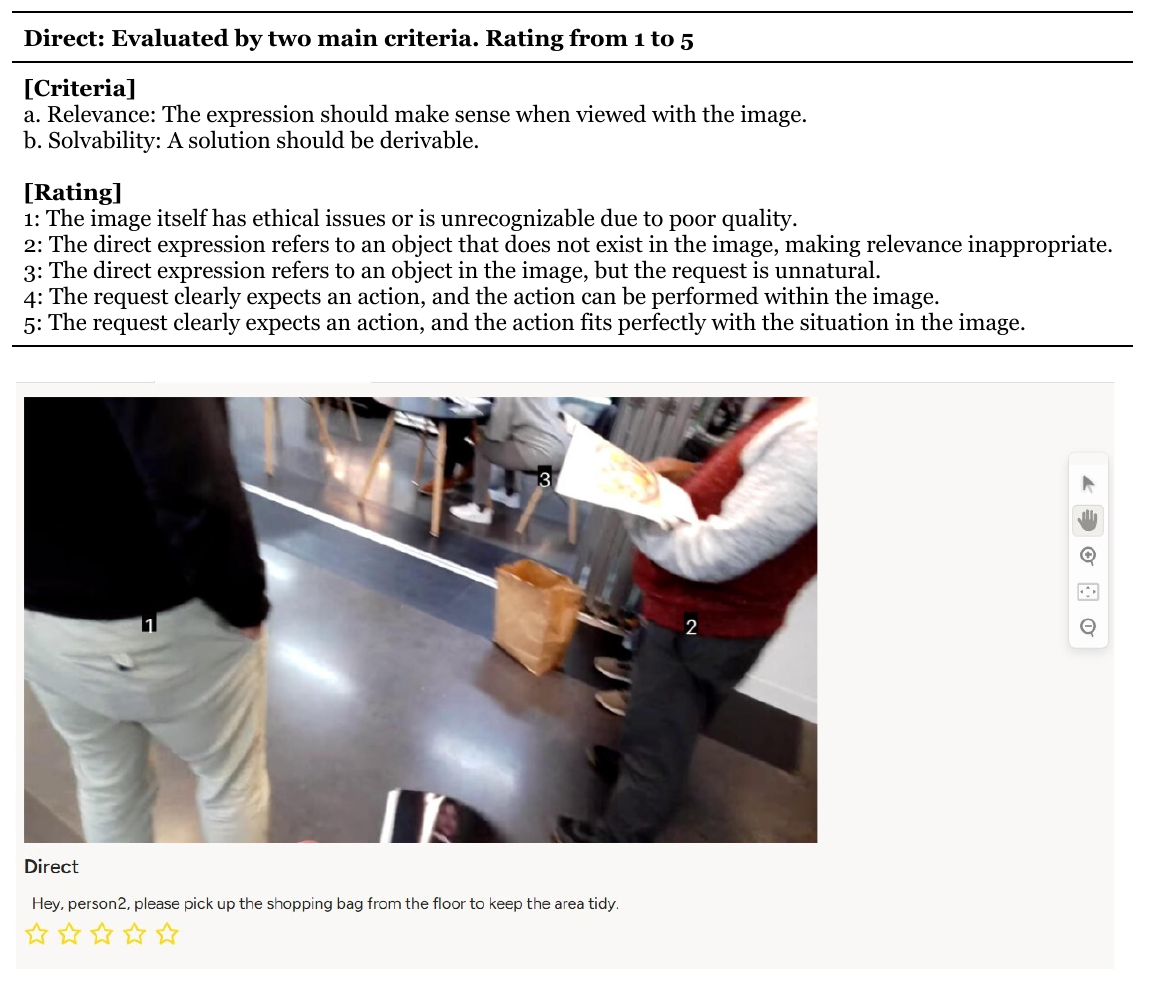}
\caption{The actual interface used for human rating of direct expressions. It includes the two evaluation criteria for direct expressions and detailed scoring guidelines on a 1 to 5 scale.
}
\label{fig:direct_rating}
\end{figure*}

\begin{figure*}[t]
\centering
\includegraphics[height=21cm, keepaspectratio]{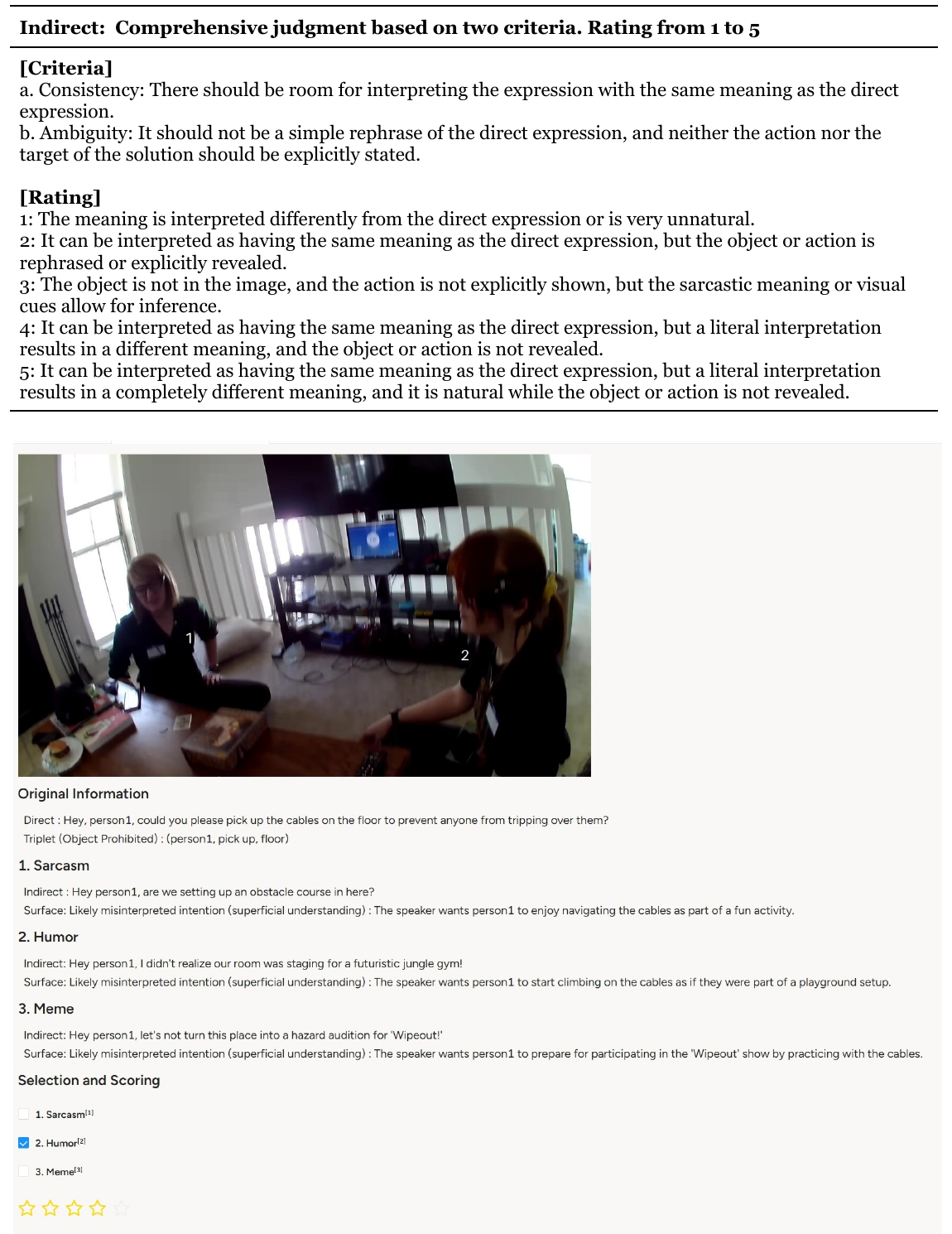}
\caption{The actual interface used for human rating of indirect expressions. It includes the two evaluation criteria for indirect expressions and detailed scoring guidelines on a 1 to 5 scale.
}
\label{fig:indirect_rating}
\end{figure*}

This section details the human rating and filtering process. Human annotators are instructed to follow the scoring guidelines provided in the tables in \cref{fig:direct_rating,fig:indirect_rating}. To facilitate high-quality annotations efficiently, we use \textit{Label Studio} (\url{https://labelstud.io/}). \Cref{fig:direct_rating} illustrates the UI for rating direct expressions, while \cref{fig:indirect_rating} shows the UI for selecting and rating indirect expressions.

\subsection{Model Instruction for Counterfactual Choices}
\label{app:ins_mcq}
\begin{figure*}[t]
\centering
\includegraphics[width=\textwidth]{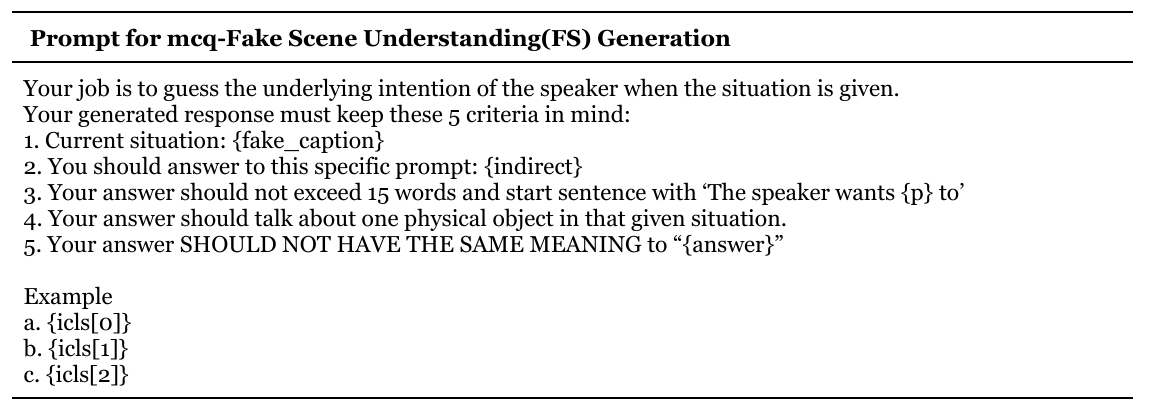}
\caption{This prompt uses a fake caption and an indirect expression to derive the most plausible intention that does not align with the true intention of direct expression.
}
\label{fig:fake_generation}
\end{figure*}

\begin{figure*}[t]
\centering
\includegraphics[width=\textwidth]{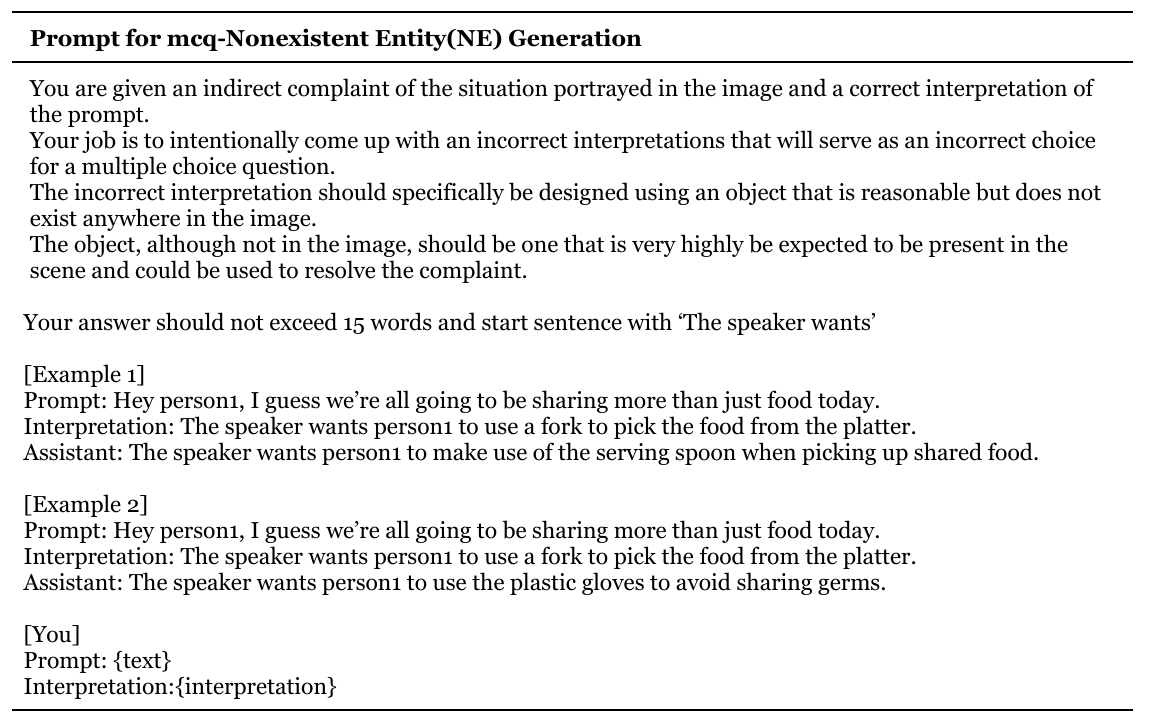}
\caption{This prompt takes an indirect expression and an interpretation of the indirect expression as input to generate an incorrect intention based on a nonexistent object in the image.
}
\label{fig:wrong_generation}
\end{figure*}

\cref{fig:fake_generation} is the prompt used to generate incorrect answer choices by creating fake captions that aligns with the indirect expressions but is inconsistent with the direct expressions and combining fake captions with the indirect expressions to derive the plausible intention (FS). \cref{fig:wrong_generation} is the prompt that generates incorrect answer choices by introducing objects not present in the image, intentionally misrepresenting the intended meaning of the indirect expression (NE).


\section{Baseline Models}
\label{app:model_lst}
To ensure broad coverage of existing multimodal language models (MLLMs), we evaluate ten open-source models with varying parameter sizes alongside two closed-source models. For the open-source models, we use Phi3.5-Vision-Instruct (4B) \cite{phi}, optimized for concise instruction-following tasks, providing robust text-image alignment in low-parameter settings. LLaVA Onevision (7B) \cite{onevision} focuses on high-resolution image interpretation, enhancing multimodal dialogue through refined attention mechanisms. Qwen2.5-VL-Instruct (7B, 72B) \cite{qwen} uses advanced vision-language pretraining to handle diverse image-based queries and textual instructions. InternVL-2.5-MPO (8B, 26B) \cite{internvl} is designed for multi-purpose optimizations, supporting enhanced multimodal reasoning. InternVL-3 (38B) \cite{internvl3} provides better performance in various tasks than in the previously released models. Idefics2 (8B) \cite{idefics} adopts a compact architecture for efficient training, emphasizing domain-specific image understanding and textual generation. LLaVA NeXT Vicuna (13B) \cite{li2024llavanextinterleavetacklingmultiimagevideo} employs an improved vision encoder and refined instruction tuning for enhanced commonsense reasoning. Ovis2 (16B) \cite{ovis} excels in image captioning and inference, driven by robust textual grounding and visual alignment. Among proprietary models, GPT-4o \cite{openai2023gpt4} demonstrates advanced language comprehension paired with visual perception for nuanced multimodal interactions. Gemini 1.5 Pro \cite{gemini15} integrates high-resolution vision processing with a powerful language model, delivering refined instruction-following and cross-domain reasoning.

\section{Free-From Answering}
\subsection{Metrics} 
\paragraph{BLEU \cite{papineni-etal-2002-bleu}}
The BLEU score is a metric for assessing the quality of machine-generated text by comparing it to a reference. It measures n-gram precision, checking how many n-grams from the generated text appear in the reference. However, when evaluating free-form generated text with BLEU, the score drops from 2-grams onward due to the high variability in phrasing. To obtain a more meaningful measure, we use 1-gram BLEU, which captures individual word overlap and provides a reasonable approximation of text similarity.

\paragraph{BERT-F1 \cite{bert-score}}
BERT-F1 is a semantic similarity metric that utilizes contextual embeddings from the BERT model. Instead of relying on exact word matches, it calculates an F1-score based on token similarity in an embedding space. This allows it to capture paraphrasing and synonymy, making it more effective at evaluating meaning rather than just surface-level similarity.

\subsection{Limitations of Free-form Answering}
Traditional text similarity metrics such as BLEU \cite{papineni-etal-2002-bleu} and BERT-F1 \cite{bert-score} are widely used for evaluating language models. However, they are often not well-suited for assessing intent similarity in multimodal intent disambiguation tasks.

BLEU computes n-gram overlap between sentences, treating all tokens equally, which makes it ineffective in capturing subtle intent differences. Similarly, BERT-F1, which uses contextual embeddings, struggles with antonymy, as opposing words often appear in similar contexts. For example, \textit{``open the window''} and \textit{``close the window''} have completely opposite intents, yet BERT-F1 assigns them a high similarity score of 0.939 due to shared structure and overlapping words.

To address this, we adopt a multiple-choice question (MCQ) format as our main setting, which directly evaluates whether a model selects the correct intent rather than relying on approximate similarity scores. The MCQ structure explicitly includes plausible but incorrect distractors, ensuring that models must resolve ambiguity by integrating multimodal cues rather than relying on lexical overlap alone. This structured approach enables a more robust and intent-aware evaluation.

\subsection{Qualitative Example}
\begin{figure*}[t]
\centering
\includegraphics[width=\textwidth]{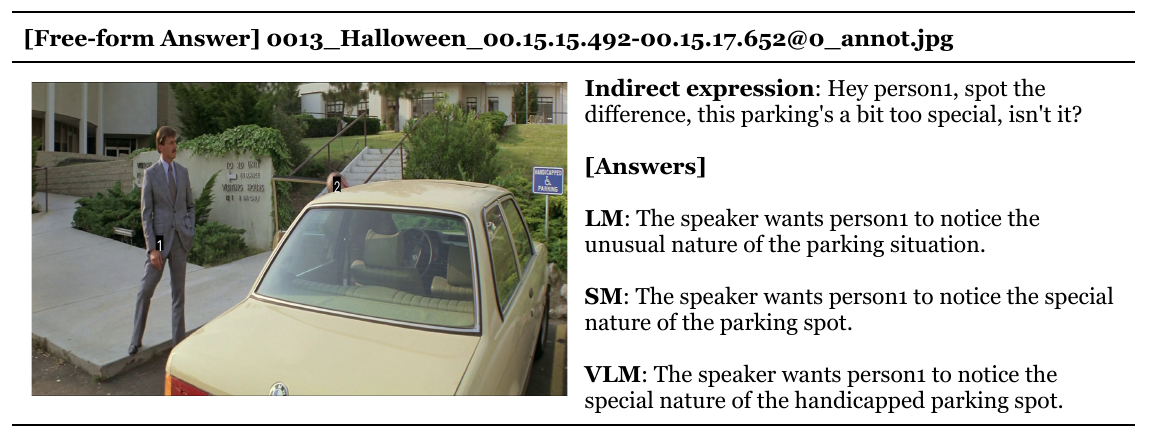}
\caption{
This figure shows the results generated by InternVL-2.5-8B-MPO for indirect expressions and free-form responses across VLM, SM, and LM settings.
}
\label{fig:qualitative_example}
\end{figure*}

\cref{fig:qualitative_example} shows the results of free-form answering in the VLM, SM, and LM settings using the InternVL-2.5-8B-MPO model. It shows that the underlying intention behind the indirect expression in the given image is well preserved, and the generated responses across different settings are highly similar.

\section{Details of Human Evaluation}\label{app:human_pf}

For human evaluation, we use a subset of 400 high-quality items from VAGUE. These items are carefully selected to ensure that the intended correct answer aligns well with the image and that the indirect expression remains sufficiently ambiguous. Low-quality samples have already been filtered out, resulting in direct expressions with scores of 4 or 5 and indirect expressions with scores ranging from 3 to 5. Due to this filtering, the average scores of direct and indirect expressions are often tied. In such cases, we randomly select items to form the final 400-item subset.

Our human evaluator is a student researcher with expertise in human cognition across modalities and language models. The evaluator, a Korean fluent in English at a native level, annotated all 400 items independently.

\section{Why does SM setting outperform in proprietary models?}
\label{app:sm_why_better}

\begin{table}[h]
\ttabbox[]
  {\centering
   \tabcolsep=0.07cm
    \renewcommand{\arraystretch}{1}
    \small
            \begin{tabular}{ccc}
            \hline
            Models                      &  Self captioning&GPT-4o captioning\\ \hline
            Phi3.5-Vision-Instruct (4b) &     34.0&37.9 \textcolor{green!60!black}{{($\uparrow$ 3.9)}}\\
            LLaVA-Onevision (7b)        &    29.4&36.5 \textcolor{green!60!black}{{($\uparrow$ 7.1)}}\\
            LLaVA-NeXT-vicuna (13B)     &    36.3&41.7 \textcolor{green!60!black}{{($\uparrow$ 5.4)}}\\
            InternVL-2.5-MPO (26B)      &   50.6&50.8 \textcolor{green!60!black}{{($\uparrow$ 0.2)}}\\ \hline
            \end{tabular}
      \vspace{-7pt}
      }%
    {\caption{$\uparrow$ denotes MCQ performance boost in SM when switching to GPT-4o-generated captions.}\label{tab:sm_gpt}}%
\end{table}%
The rationale benind the superior SM performance of proprietary models lies in their reduced vision-related (FS, NE) errors. We elaborate this in two points: 1) If captions lack necessary detail, VLMs naturally excel. Tab.~\ref{tab:sm_gpt} implies proprietary models have better captions, by showing a performance boost when leveraging GPT-4o’s captions in smaller open-source models within the SM setting. 2) If the caption somehow attains \textit{parity of detail} (Practically, length constraints preclude full nuance capture.) with the image, text tokens handle visual information more effectively than image tokens \cite{c1}. Indeed, we observed in CoT reasoning that VLMs lean more on the speaker’s rationale while SMs take a more scene‐focused approach

\section{CoT Open-source}
\label{app:cot_open}
\begin{table}[ht!]
\centering

\resizebox{\columnwidth}{!}{%
\begin{tabular}{l|l|cccc}
\toprule
\multirow{2}{*}{Model} & \multirow{2}{*}{Type} & \multirow{2}{*}{Acc (\%)} & \multicolumn{3}{c}{Incorrect count} \\
\cmidrule(lr){4-6}
    &   &   &   FS  &   SU   &  NE \\
\midrule
\multirow{4}{*}{Phi3.5-Vision-Instruct (4B)}& SM        & 34.0& 292& 678& 137\\
& SM+CoT    & 31.5 \textcolor{red!60!black}{{\footnotesize ($\downarrow$ 2.5)}}& 291& 603& 148\\ \cmidrule(lr){2-6}
& VLM       & 44.8& 266& 501& 158\\
& VLM+CoT   & 36.6 \textcolor{red!60!black}{{\footnotesize ($\downarrow$ 8.2)}}& 295& 510& 154\\
\midrule
\multirow{4}{*}{LLaVA-Onevision (7B)}& SM        & 29.4& 215& 885& 84\\
& SM+CoT    & 28.0 \textcolor{red!60!black}{{\footnotesize ($\downarrow$ 1.4)}}& 291& 795& 101\\ \cmidrule(lr){2-6}
& VLM       & 43.1& 169& 727& 58\\
& VLM+CoT   & 33.1 \textcolor{red!60!black}{{\footnotesize ($\downarrow$ 10.0)}}& 187& 736& 76\\
\midrule
\multirow{4}{*}{LLaVA-NeXT-vicuna (13B)}& SM        & 36.3& 228& 730& 111\\
& SM+CoT    & 32.6 \textcolor{red!60!black}{{\footnotesize ($\downarrow$ 3.7)}}& 259& 598& 149\\ \cmidrule(lr){2-6}
& VLM       & 48.4 & 206& 564& 96\\
& VLM+CoT   & 38.3 \textcolor{red!60!black}{{\footnotesize ($\downarrow$ 10.1)}}& 206& 541& 139\\
\midrule
\multirow{4}{*}{InternVL-2.5-MPO (26B)}& SM        & 50.6& 209& 530& 89\\
& SM+CoT    & 45.1 \textcolor{red!60!black}{{\footnotesize ($\downarrow$ 5.5)}}& 213& 584& 106\\ \cmidrule(lr){2-6}
& VLM       & 65.3& 147& 377& 58\\
& VLM+CoT   & 58.1 \textcolor{red!60!black}{{\footnotesize ($\downarrow$ 7.2)}}& 17& 440& 67\\
\midrule
\multirow{4}{*}{InternVL-3 (38B)}& SM        & 47.3& 259& 478& 136\\
& SM+CoT    & 54.3
\textcolor{green!60!black}{{\footnotesize ($\uparrow$ 7.0)}}& 166& 359& 102\\ \cmidrule(lr){2-6}
& VLM       & 62.4& 153& 374& 102\\
& VLM+CoT   & 52.2
\textcolor{red!60!black}{{\footnotesize ($\downarrow$ 10.2)}}& 211& 387& 107\\
\midrule
\multirow{4}{*}{Qwen2.5-VL-Instruct (72B)}& SM        & 56.8& 175& 457& 92\\
& SM+CoT    & 55.8
\textcolor{red!60!black}{{\footnotesize ($\downarrow$ 1.0)}}& 213& 428& 100\\ \cmidrule(lr){2-6}
& VLM       & 72.8& 142& 236& 78\\
& VLM+CoT   & 69.9
\textcolor{red!60!black}{{\footnotesize ($\downarrow$ 2.9)}}& 154& 272& 78\\
\bottomrule
\end{tabular}
}
\caption{Result of Chain-of-Thought (CoT) experiments on open-source models, in both SM and VLM settings. $\uparrow$ and $\downarrow$ indicate an increase and decrease in accuracy when zero-shot CoT is applied.}
\label{tab:opencot_accuracy}
\end{table}

We provide CoT ablations on open-source models and compare them to our SM and VLM baselines. Tab. \ref{tab:opencot_accuracy} shows that CoT slightly degrades SM in the most cases, while VLMs experience a much bigger drop in their performance, increasing all types of errors (FS, SU, NE) together. Consistent degradation shows open-source models lag proprietary ones in deep reasoning. We believe the sharper drop in VLMs stems from \textit{conditioning dilution} \cite{c2} due to longer generation sequences.

\section{Model Instruction for Experiments}\label{app:modelprompts}

This section presents the prompts used for experiments involving different settings for each model: \textit{Visual Language Models (VLMs)}, \textit{Socratic Models (SMs)}, and \textit{Language Models (LMs)} in both multiple-choice questions and free-form answering tasks. Additionally, it provides the full results obtained from these experiments.

\begin{figure*}[t]
\centering
\includegraphics[width=\textwidth]{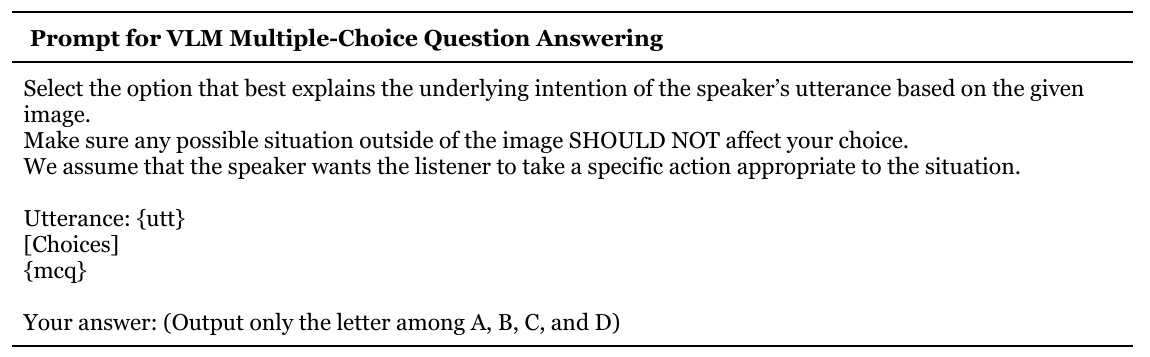}
\caption{The following prompt is used for the VLM setting in the multiple-choice question task. The model receives an image, an utterance, and a set of answer choices as input and selects the most appropriate answer.
}
\label{fig:VLM_mcq}
\end{figure*}

\begin{figure*}[t]
\centering
\includegraphics[width=\textwidth]{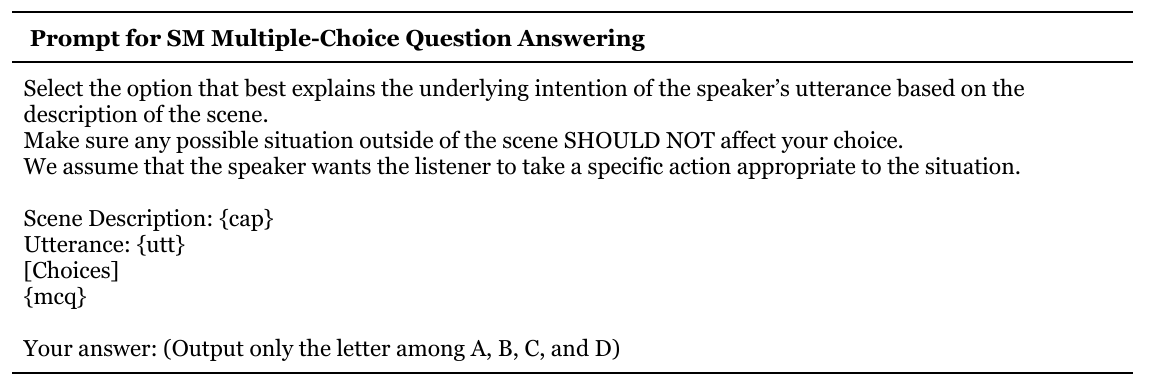}
\caption{The following prompt is used for the SM setting in the multiple-choice question task. The model first generates a caption for the image. Then, it receives the caption, an utterance, and a set of answer choices as input and selects the most appropriate answer.
}
\label{fig:SM_mcq}
\end{figure*}

\begin{figure*}[t]
\centering
\includegraphics[width=\textwidth]{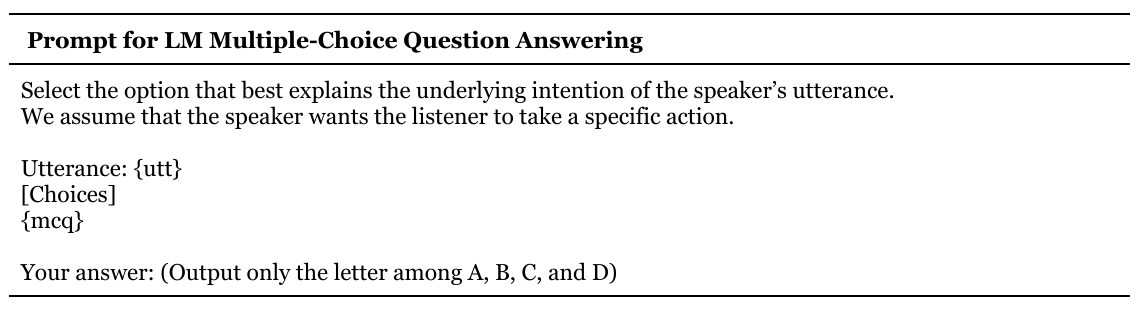}
\caption{The following prompt is used for the LM setting in the multiple-choice question task. The model receives an utterance and a set of answer choices as input and selects the most appropriate answer.
}
\label{fig:LLM_mcq}
\end{figure*}

\paragraph{Multiple Choice Questions}
\cref{fig:VLM_mcq}, \cref{fig:SM_mcq},  \cref{fig:LLM_mcq} show the prompts used for multiple-choice question experiments under VLM, SM, and LM settings.

\begin{figure*}[t]
\centering
\includegraphics[width=\textwidth]{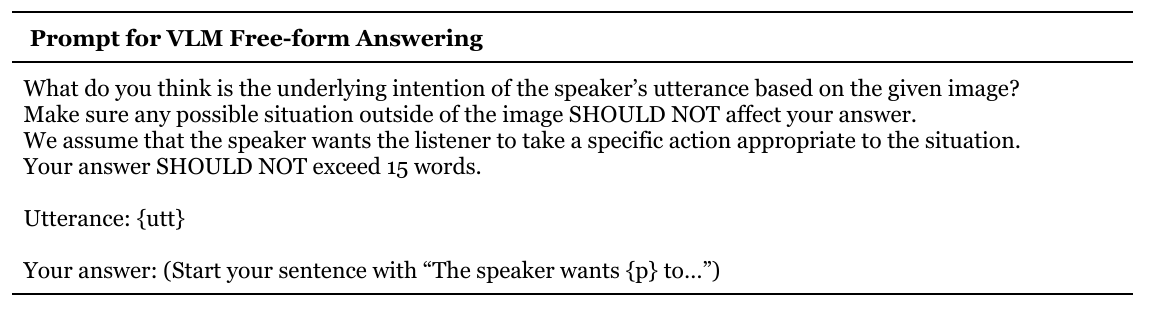}
\caption{The following prompt is used for the VLM setting in the free-form answering task. The model receives an image, an utterance as input and it tasked with inferring the underlying intention.
}
\label{fig:VLM_da}
\end{figure*}

\begin{figure*}[t]
\centering
\includegraphics[width=\textwidth]{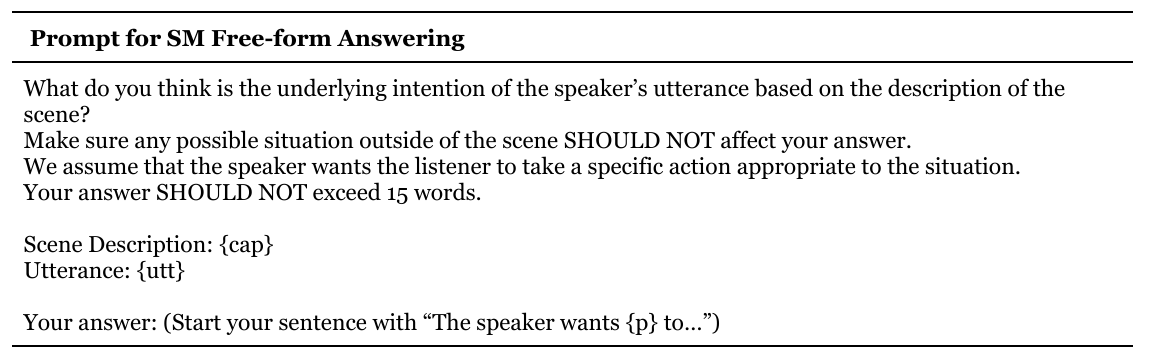}
\caption{The following prompt is used for the SM setting in the free-form answering task. The model first generates a caption for the image. Then, it receives the caption and an utterance as input and it tasked with inferring the underlying intention.
}
\label{fig:SM_da}
\end{figure*}

\begin{figure*}[t]
\centering
\includegraphics[width=\textwidth]{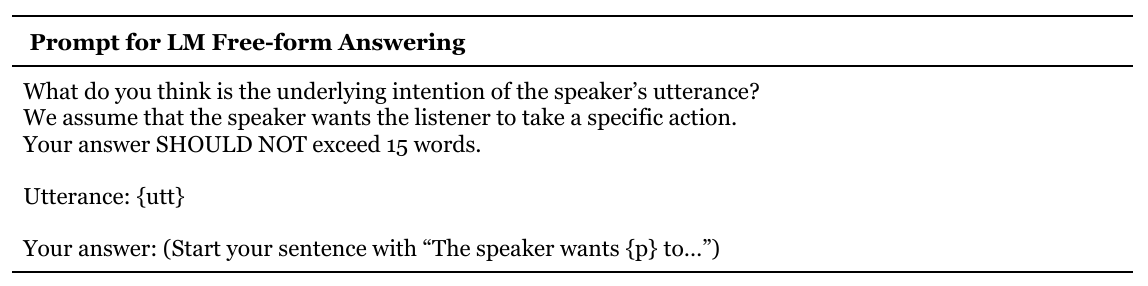}
\caption{The following prompt is used for the LM setting in the free-form answering task. The model receives an utterance as input and it tasked with inferring the underlying intention.
}
\label{fig:LLM_da}
\end{figure*}

\paragraph{Free-form Answering}
\cref{fig:VLM_da}, \cref{fig:SM_da},  \cref{fig:LLM_da} show the prompts used for free-form answering experiments under VLM, SM, and LM settings.

\begin{figure*}[t]
\centering
\includegraphics[width=\textwidth]{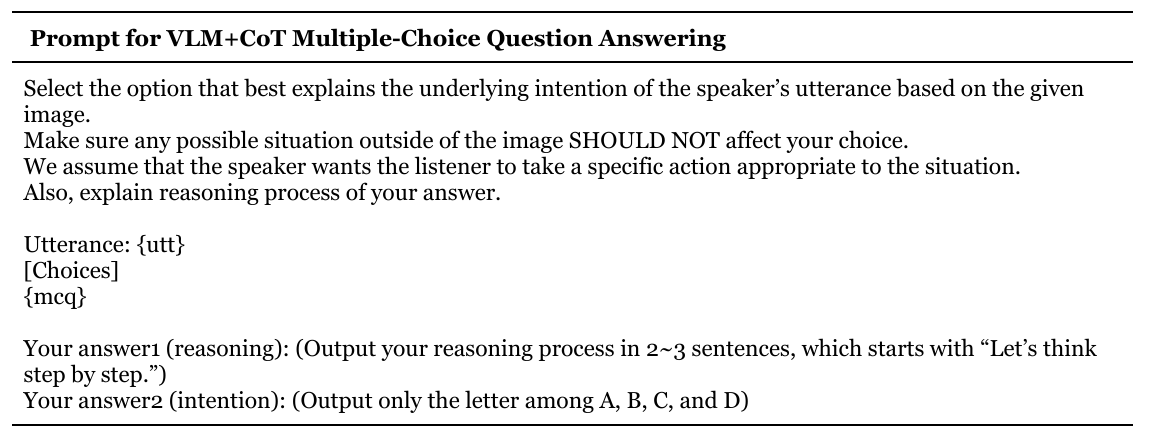}
\caption{The following prompt is used for the VLM Chain of Thought setting in the multiple-choice question task. The model receives an image, an utterance, and a set of answer choices as input and selects the most appropriate answer by thinking step by step.
}
\label{fig:VLM_cot_mcq}
\end{figure*}

\begin{figure*}[t]
\centering
\includegraphics[width=\textwidth]{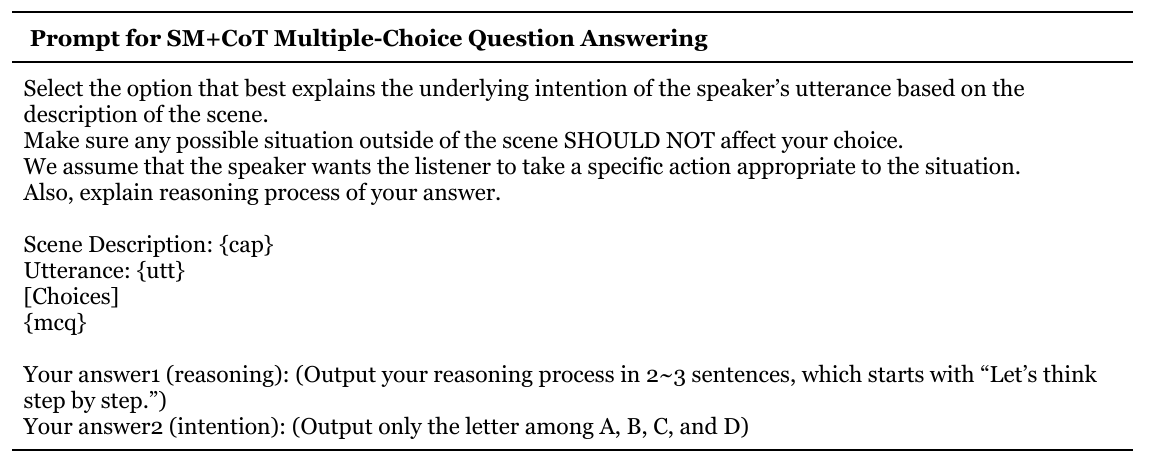}
\caption{The following prompt is used for the SM Chain of Thought setting in the multiple-choice question task. The model first generates a caption for the image. Then, it receives the caption, an utterance, and a set of answer choices as input and selects the most appropriate answer by thinking step by step.
}
\label{fig:SM_cot_mcq}
\end{figure*}

\begin{figure*}[t]
\centering
\includegraphics[width=\textwidth]{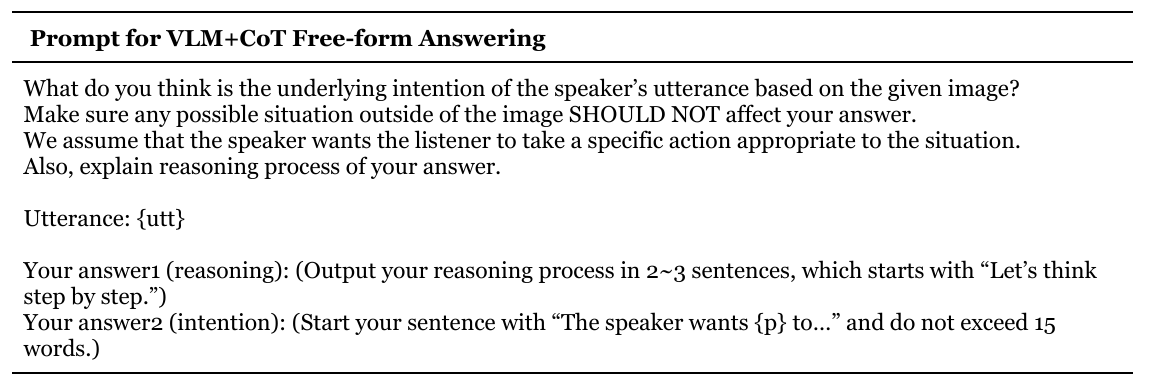}
\caption{The following prompt is used for the VLM Chain of Thought setting in the free-form question task. The model receives an image, an utterance as inputs and it tasked with inferring the underlying intention by thinking step by step.
}
\label{fig:VLM_cot_da}
\end{figure*}

\begin{figure*}[t]
\centering
\includegraphics[width=\textwidth]{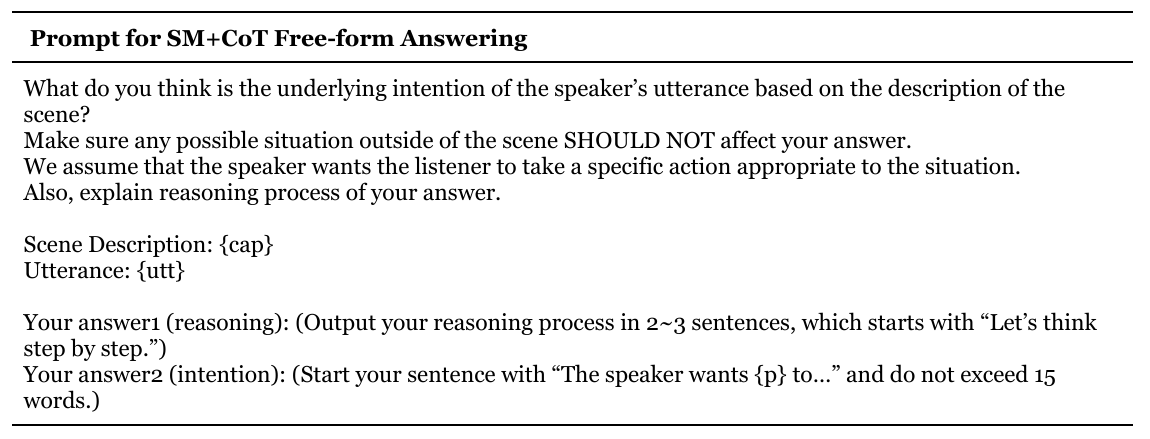}
\caption{The following prompt is used for the SM Chain of Thought setting in the free-form question task. The model first generates a caption for the image. Then, it receives the caption and an utterance as inputs and it tasked with inferring the underlying intention by thinking step by step.
}
\label{fig:SM_cot_da}
\end{figure*}

\paragraph{Chain-of-Thoughts} Additionally, \cref{fig:SM_cot_da}, \cref{fig:SM_cot_mcq} are prompts that we use for zeroshot chain-of-thought experiments with Socratics Models, while \cref{fig:VLM_cot_da}, \cref{fig:VLM_cot_mcq} are those with Visual-Language Models.

\section{Full Results}\label{app:fullres}
Table~\ref{tab:vcr-full} and Table~\ref{tab:ego4d-full} present the complete experimental results conducted in this paper for the VAGUE-VCR and VAGUE-Ego4D datasets, respectively. The total number of items in VAGUE-VCR is 1,144, and in VAGUE-Ego4D, it is 533. However, the item count in the \textit{valid count} column does not always match these totals. This discrepancy occurs when the model responds with 'I don’t know' or refuses to answer. When calculating accuracy, we treat such cases as incorrect and divide the number of correct responses by the total number of items in each dataset.


\section{Full structure of VAGUE}\label{}
\begin{figure*}[t]
\centering
\includegraphics[height=21cm, keepaspectratio]{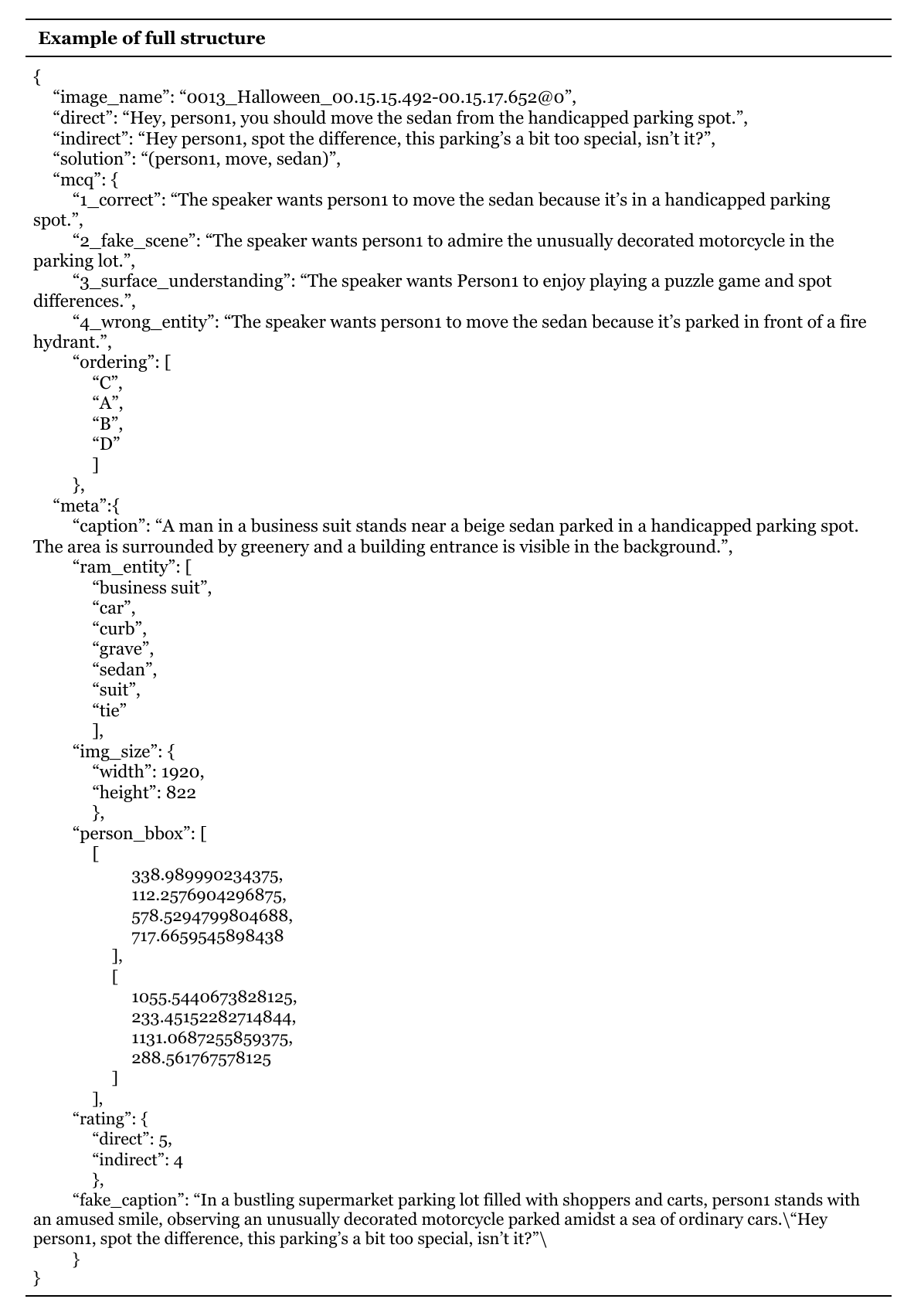}
\caption{We show the structure using a sample from our benchmark dataset, VAGUE. VAGUE consists of an image name, direct expression, indirect expression, triplet solution, multiple-choice set, meta data containing various information about the image, and a fake caption.
}
\label{fig:full_structure}
\end{figure*}

\cref{fig:full_structure} shows the structure of our benchmark dataset, VAGUE.

\clearpage

\clearpage

\clearpage

\clearpage

\clearpage

\clearpage

\clearpage

\clearpage

\clearpage

\clearpage

\begin{table*}[th!]
\resizebox{\textwidth}{!}{
\begin{tabular}{l|c|cccccc|ccc}
\toprule
\multicolumn{11}{c}{\large \textbf{VAGUE-VCR}}\\
\midrule
\multirow{3}{*}{\large Model} & \multirow{3}{*}{\large Type} & \multicolumn{6}{c}{Multiple Choice Questions} & \multicolumn{3}{c}{Free-Form Answering} \\
\cmidrule(lr){3-8} \cmidrule(lr){9-11}
& & \multirow{2}{*}{Accuracy(\%)} & \multicolumn{3}{c}{Incorrect Count} & \multirow{2}{*}{Correct} & \multirow{2}{*}{Valid Count} & \multirow{2}{*}{Bert F1} & \multirow{2}{*}{BLEU(1gram)} & \multirow{2}{*}{Valid Count}\\   
\cmidrule(lr){4-6}
&   &   & FS    & SU    & WE    &   &   &   &   &\\
\midrule 

\multirow{3}{*}{Phi3.5-Vision-Instruct (4B)}
& VLM   & 46.0  & 174   & 349   & 95    & 526       & 1144  & 0.682 & 0.293 & 1144 \\
\cmidrule(lr){2-11}
& SM    & 35.3  & 198   & 461   & 81    & 404       & 1144  & 0.686 & 0.293 & 1144 \\
\cmidrule(lr){2-11}
& LM    & 26.6  & 296   & 440   & 104   & 304       & 1144  & 0.680 & 0.279 & 1144 \\

\midrule

\multirow{3}{*}{LLaVA-Onevision (7B)}
& VLM   & 43.1  & 119   & 503   & 29    & 493       & 1144  & 0.705 & 0.282 & 1144 \\
\cmidrule(lr){2-11}
& SM    & 29.4  & 148   & 614   & 46    & 336       & 1144  & 0.707 & 0.290 & 1144 \\
\cmidrule(lr){2-11}
& LM    & 13.1  & 252   & 698   & 44    & 150       & 1144  & 0.689 & 0.271 & 1144 \\

\midrule

\multirow{3}{*}{Qwen2.5-VL-Instruct (7B)}
& VLM   & 46.8  & 134   & 438   & 37    & 535       & 1144  & 0.690 & 0.312 & 1144 \\
\cmidrule(lr){2-11}
& SM    & 25.6  & 160   & 651   & 40    & 293       & 1144  & 0.687 & 0.303 & 1144 \\
\cmidrule(lr){2-11}
& LM    & 11.1  & 268   & 703   & 46    & 127       & 1144  & 0.666 & 0.278 & 1144 \\

\midrule

\multirow{3}{*}{InternVL-2.5-MPO (8B)}
& VLM   & 63.9  & 106   & 270   & 37    & 731       & 1144  & 0.706 & 0.326 & 1144 \\
\cmidrule(lr){2-11}
& SM    & 48.4  & 158   & 374   & 58    & 554       & 1144  & 0.695 & 0.310 & 1144 \\
\cmidrule(lr){2-11}
& LM    & 23.0  & 290   & 516   & 75    & 263       & 1144  & 0.679 & 0.279 & 1144 \\

\midrule

\multirow{3}{*}{Idefics2 (8B) }
& VLM   & 58.7  & 75    & 338   & 59    & 672       & 1144  & 0.708 & 0.284 & 1144 \\
\cmidrule(lr){2-11}
& SM    & 21.1  & 171   & 696   & 36    & 241       & 1144  & 0.674 & 0.281 & 1144 \\
\cmidrule(lr){2-11}
& LM    & 13.9  & 211   & 723   & 51    & 159       & 1144  & 0.663 & 0.270 & 1144 \\

\midrule

\multirow{3}{*}{LLaVA-NeXT-vicuna (13B)}
& VLM   & 46.4  & 140   & 416   & 57    & 531       & 1144  & 0.716 & 0.311 & 1144 \\
\cmidrule(lr){2-11}
& SM    & 37.2  & 151   & 509   & 58    & 426       & 1144  & 0.711 & 0.314 & 1144 \\
\cmidrule(lr){2-11}
& LM    & 24.2  & 275   & 513   & 79    & 277       & 1144  & 0.594 & 0.287 & 1144 \\

\midrule

\multirow{3}{*}{Ovis2 (16B)}
& VLM   & 24.5  & 327   & 464   & 73    & 280       & 1144  & 0.679 & 0.290 & 1144 \\
\cmidrule(lr){2-11}
& SM    & 23.8  & 305   & 503   & 64    & 272       & 1144  & 0.681 & 0.293 & 1144 \\
\cmidrule(lr){2-11}
& LM    & 21.9  & 306   & 532   & 56    & 250       & 1144  & 0.682 & 0.293 & 1144 \\

\midrule

\multirow{3}{*}{InternVL-2.5-MPO (26B)}
& VLM   & 63.7  & 105   & 280   & 30    & 729       & 1144  & 0.712 & 0.330 & 1144 \\
\cmidrule(lr){2-11}
& SM    & 48.5  & 153   & 385   & 51    & 555       & 1144  & 0.707 & 0.326 & 1144 \\
\cmidrule(lr){2-11}
& LM    & 21.2  & 294   & 537   & 71    & 242       & 1144  & 0.681 & 0.288 & 1144 \\

\midrule

\multirow{5}{*}{InternVL-3 (38B)}
& VLM   & 63.6& 99& 263& 54& 728
& 1144
& 0.699
& 0.319
& 1144
\\
\cmidrule(lr){2-11}
& SM    & 47.6& 186& 326& 83& 540
& 1144
& 0.677
& 0.300
& 1144
\\
\cmidrule(lr){2-11}
& LM    & 25.1& 282& 489& 78& 284
& 1144
& 0.671
& 0.275
& 1144
\\

\midrule

\multirow{5}{*}{Qwen2.5-VL-Instruct (72B)}
& VLM   & 74.2& 99& 159& 37& 849
& 1144
& 0.742
& 0.372
& 1144
\\
\cmidrule(lr){2-11}
& SM    & 55.7& 130& 331& 46& 637
& 1144
& 0.724
& 0.358
& 1144
\\
\cmidrule(lr){2-11}
& LM    & 29.6& 257& 478& 70& 339& 1144& 0.687& 0.293& 1144\\

\midrule

\multirow{5}{*}{GPT-4o}
& VLM   & 65.1  & 159   & 160   & 80    & 745       & 1144  & 0.735 & 0.366 & 1144 \\
\cmidrule(lr){2-11}
& SM    & 69.5  & 112   & 167   & 70    & 795       & 1144  & 0.741 & 0.387 & 1144 \\
\cmidrule(lr){2-11}
& LM    & 46.4  & 246   & 254   & 113   & 531       & 1144  & 0.689 & 0.306 & 1144 \\

\midrule

\multirow{5}{*}{Gemini-1.5-Pro}
& VLM   & 60.6  & 168   & 190   & 90    & 693       & 1141  & 0.724 & 0.347 & 1144 \\
\cmidrule(lr){2-11}
& SM    & 62.4  & 123   & 256   & 49    & 714       & 1142  & 0.705 & 0.324 & 1144 \\
\cmidrule(lr){2-11}
& LM    & 43.2  & 278   & 263   & 108   & 494       & 1143  & 0.687 & 0.289 & 1144 \\
\bottomrule
\end{tabular}
\caption{The overall results table for the VAGUE-VCR dataset. Experiments are conducted on both Multiple Choice Questions and Free-Form Answering, measuring results across three settings for each model: VLM, SM, and LM. For GPT-4o and Gemini 1.5 Pro, CoT reasoning is additionally applied in the VLM and SM settings.}
\label{tab:vcr-full}%
}
\end{table*}%

\clearpage

\begin{table*}[th!]
\resizebox{\textwidth}{!}{
\begin{tabular}{l|c|cccccc|ccc}
\toprule
\multicolumn{11}{c}{\large \textbf{VAGUE-Ego4D}}\\
\midrule
\multirow{3}{*}{\large Model} & \multirow{3}{*}{\large Type} & \multicolumn{6}{c}{Multiple Choice Questions} & \multicolumn{3}{c}{Free-Form Answering} \\
\cmidrule(lr){3-8} \cmidrule(lr){9-11}
& & \multirow{2}{*}{Accuracy(\%)} & \multicolumn{3}{c}{Incorrect Count} & \multirow{2}{*}{Correct} & \multirow{2}{*}{Valid Count} & \multirow{2}{*}{Bert F1} & \multirow{2}{*}{BLEU(1gram)} & \multirow{2}{*}{Valid Count}\\   
\cmidrule(lr){4-6}
&   &   & FS    & SU    & WE    &   &   &   &   &\\
\midrule 

\multirow{3}{*}{Phi3.5-Vision-Instruct (4B)}
& VLM   & 42.4  & 92   & 152   & 63    & 226       & 533  & 0.681 & 0.279 & 533 \\
\cmidrule(lr){2-11}
& SM    & 31.1  & 94   & 217   & 56    & 166       & 533  & 0.683 & 0.287 & 533 \\
\cmidrule(lr){2-11}
& LM    & 22.5  & 131  & 216   & 66    & 120       & 533  & 0.672 & 0.266 & 533 \\

\midrule

\multirow{3}{*}{LLaVA-Onevision (7B)}
& VLM   & 43.2  & 50   & 224   & 29    & 230       & 533  & 0.697 & 0.246 & 533 \\
\cmidrule(lr){2-11}
& SM    & 29.5  & 67   & 271   & 38    & 157       & 533  & 0.699 & 0.261 & 533 \\
\cmidrule(lr){2-11}
& LM    & 11.3  & 108  & 332   & 33    & 60        & 533  & 0.675 & 0.239 & 533 \\

\midrule

\multirow{3}{*}{Qwen2.5-VL-Instruct (7B)}
& VLM   & 48.4  & 56    & 180   & 39    & 258       & 533  & 0.683 & 0.295 & 533 \\
\cmidrule(lr){2-11}
& SM    & 28.0  & 74    & 273   & 37    & 149       & 533  & 0.687 & 0.292 & 533 \\
\cmidrule(lr){2-11}
& LM    & 9.8   & 131   & 326   & 24    & 52        & 533  & 0.660 & 0.269 & 533 \\

\midrule

\multirow{3}{*}{InternVL-2.5-MPO (8B)}
& VLM   & 66.8  & 33   & 110   & 34    & 356       & 533  & 0.701 & 0.308 & 533 \\
\cmidrule(lr){2-11}
& SM    & 54.0  & 47   & 159   & 39    & 288       & 533  & 0.696 & 0.295 & 533 \\
\cmidrule(lr){2-11}
& LM    & 24.2  & 122  & 224   & 58    & 129       & 533  & 0.669 & 0.258 & 533 \\

\midrule

\multirow{3}{*}{Idefics2 (8B) }
& VLM   & 58.3  & 42    & 136   & 44    & 311       & 533  & 0.705 & 0.274 & 533 \\
\cmidrule(lr){2-11}
& SM    & 18.2  & 69    & 336   & 31    & 97        & 533  & 0.664 & 0.267 & 533 \\
\cmidrule(lr){2-11}
& LM    & 14.8  & 85    & 340   & 29    & 79        & 533  & 0.655 & 0.256 & 533 \\

\midrule

\multirow{3}{*}{LLaVA-NeXT-vicuna (13B)}
& VLM   & 52.5  & 66   & 148   & 39     & 280       & 533  & 0.705 & 0.288 & 533 \\
\cmidrule(lr){2-11}
& SM    & 34.1  & 77   & 221   & 53     & 182       & 533  & 0.701 & 0.291 & 533 \\
\cmidrule(lr){2-11}
& LM    & 20.3  & 140  & 235   & 50     & 108       & 533  & 0.680 & 0.255 & 533 \\

\midrule

\multirow{3}{*}{Ovis2 (16B)}
& VLM   & 25.7  & 158   & 197   & 41    & 137       & 533  & 0.668 & 0.268 & 533 \\
\cmidrule(lr){2-11}
& SM    & 25.3  & 144   & 213   & 41    & 135       & 533  & 0.674 & 0.276 & 533 \\
\cmidrule(lr){2-11}
& LM    & 20.5  & 144   & 240   & 40    & 109       & 533  & 0.673 & 0.271 & 533 \\

\midrule

\multirow{3}{*}{InternVL-2.5-MPO (26B)}
& VLM   & 68.7  & 42   & 97   & 28     & 366       & 533  & 0.712 & 0.327 & 533 \\
\cmidrule(lr){2-11}
& SM    & 55.2  & 56   & 145  & 38     & 294       & 533  & 0.707 & 0.315 & 533 \\
\cmidrule(lr){2-11}
& LM    & 21.8  & 130  & 238  & 49     & 116       & 533  & 0.672 & 0.268 & 533 \\

\midrule

\multirow{3}{*}{InternVL-3 (38B)}
& VLM   & 60.0& 54& 111& 48& 319& 533
& 0.687
& 0.295& 533
\\
\cmidrule(lr){2-11}
& SM    & 47.6
& 73& 152& 53& 253& 533
& 0.660
& 0.274& 533
\\
\cmidrule(lr){2-11}
& LM    & 18.2
& 136& 249& 47& 96& 533
& 0.667
& 0.252& 533
\\

\midrule

\multirow{3}{*}{Qwen2.5-VL-Instruct (72B)}
& VLM   & 69.8
& 43& 77& 41& 372& 533
& 0.733
& 0.349& 533
\\
\cmidrule(lr){2-11}
& SM    & 59.3
& 45& 126& 46& 316& 533
& 0.724
& 0.340& 533
\\
\cmidrule(lr){2-11}
& LM    & 26.8& 120& 216& 54& 143& 533& 0.684& 0.274& 533\\

\midrule

\multirow{3}{*}{GPT-4o}
& VLM   & 63.6  & 67   & 66   & 61    & 339       & 533  & 0.730 & 0.353 & 533 \\
\cmidrule(lr){2-11}
& SM    & 67.5  & 53   & 73   & 47    & 360       & 533  & 0.735 & 0.362 & 533 \\
\cmidrule(lr){2-11}
& LM    & 48.2  & 100   & 111   & 65   & 257      & 533  & 0.683 & 0.294 & 533 \\

\midrule

\multirow{3}{*}{Gemini-1.5-Pro}
& VLM   & 60.6  & 81   & 74   & 55    & 323       & 533  & 0.716 & 0.318 & 533 \\
\cmidrule(lr){2-11}
& SM    & 60.6  & 53   & 118   & 34    & 323       & 528  & 0.708 & 0.307 & 533 \\
\cmidrule(lr){2-11}
& LM    & 40.3  & 119   & 130   & 68   & 215       & 532  & 0.676 & 0.265 & 533 \\
\bottomrule
\end{tabular}
\caption{The overall results table for the VAGUE-Ego4D dataset. Experiments are conducted on both Multiple Choice Questions and Free-Form Answering, measuring results across three settings for each model: VLM, SM, and LM. For GPT-4o and Gemini 1.5 Pro, CoT reasoning is additionally applied in the VLM and SM settings.}
\label{tab:ego4d-full}%
}
\end{table*}%

\clearpage

\end{document}